\documentclass[10pt,twocolumn,letterpaper]{article}
    \makeatletter
\def\@fnsymbol#1{\ensuremath{\ifcase#1\or \dagger\or \ddagger\or
   \mathsection\or \mathparagraph\or \|\or **\or \dagger\dagger
   \or \ddagger\ddagger \else\@ctrerr\fi}}
    \makeatother
    
\usepackage{cvpr}              

\usepackage{graphicx}
\usepackage{amsmath}
\usepackage{amssymb}
\usepackage{booktabs}
\usepackage{multirow}
\usepackage{cases}

\usepackage[pagebackref,breaklinks,colorlinks]{hyperref}

\usepackage[capitalize]{cleveref}
\crefname{section}{Sec.}{Secs.}
\Crefname{section}{Section}{Sections}
\Crefname{table}{Table}{Tables}
\crefname{table}{Tab.}{Tabs.}

\def\cvprPaperID{5271} 
\def\confName{CVPR}
\def\confYear{2023}

\def \mbb{\mathbb}
\def \tbf{\textbf}

\begin{document}

\title{Natural Language-Assisted Sign Language Recognition}

\author{
Ronglai Zuo\textsuperscript{\rm 1} \qquad Fangyun Wei\textsuperscript{\rm 2}\footnotemark[1]\thanks{Corresponding author.} \qquad Brian Mak\textsuperscript{\rm 1}\\
\textsuperscript{\rm 1}The Hong Kong University of Science and Technology \quad \textsuperscript{\rm 2}Microsoft Research Asia\\
\texttt{\small\{rzuo,mak\}@cse.ust.hk} \qquad
\texttt{\small fawe@microsoft.com} 
}
\maketitle

\begin{abstract}
Sign languages are visual languages which convey information by signers' handshape, facial expression, body movement, and so forth. Due to the inherent restriction of combinations of these visual ingredients, there exist a significant number of visually indistinguishable signs (VISigns) in sign languages, which limits the recognition capacity of vision neural networks.
To mitigate the problem, we propose the Natural Language-Assisted Sign Language Recognition (NLA-SLR) framework, which exploits semantic information contained in glosses (sign labels). First, for VISigns with similar semantic meanings, we propose language-aware label smoothing by generating soft labels for each training sign whose smoothing weights are computed from the normalized semantic similarities among the glosses to ease training. Second, for VISigns with distinct semantic meanings, we present an inter-modality mixup technique which blends vision and gloss features to further maximize the separability of different signs under the supervision of blended labels. Besides, we also introduce a novel backbone, video-keypoint network, which not only models both RGB videos and human body keypoints but also derives knowledge from sign videos of different temporal receptive fields. Empirically, our method achieves state-of-the-art performance on three widely-adopted benchmarks: MSASL, WLASL, and NMFs-CSL. Codes are available at \href{https://github.com/FangyunWei/SLRT}{https://github.com/FangyunWei/SLRT}.
\end{abstract}
\vspace{-5mm}
\section{Introduction}
\label{sec:intro}
Sign languages are the primary languages for communication among deaf communities. On the one hand, sign languages have their own linguistic properties as most natural languages \cite{sandler2006sign, adaloglou2021comprehensive, yin2022mlslt}. On the other hand, sign languages are visual languages that convey information by the movements of the hands, body, head, mouth, and eyes, making them completely separate and distinct from natural languages \cite{stmc, zuo2022c2slr, chen2022simple}. This work dedicates to sign language recognition (SLR), which requires models to classify the isolated signs from videos into a set of glosses\footnote{Gloss is a unique label for a single sign. Each gloss is identified by a word which is associated with the sign’s semantic meaning.}. Despite its fundamental capacity of recognizing signs, SLR has a broad range of applications including sign spotting~\cite{varol2022scaling, momeni2022automatic, li2020transferring}, sign video retrieval~\cite{duarte2022sign}, sign language translation~\cite{li2020tspnet, shi2022open, chen2022simple}, and continuous sign language recognition~\cite{chen2022simple, adaloglou2021comprehensive}.

\begin{figure}
     \centering
     \begin{subfigure}[b]{0.45\textwidth}
         \centering
         \includegraphics[width=\textwidth]{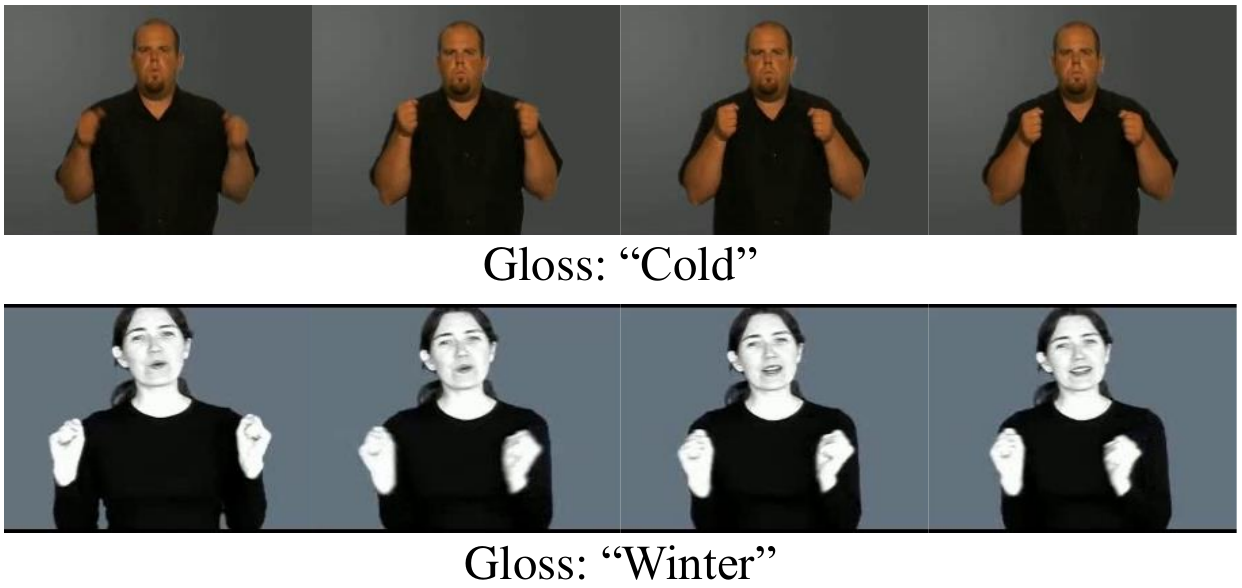}
         \caption{VISigns may have \textit{similar} semantic meanings.}
         \label{fig:teaser_A}
     \end{subfigure}
     \hfill
     \begin{subfigure}[b]{0.45\textwidth}
         \centering
         \includegraphics[width=\textwidth]{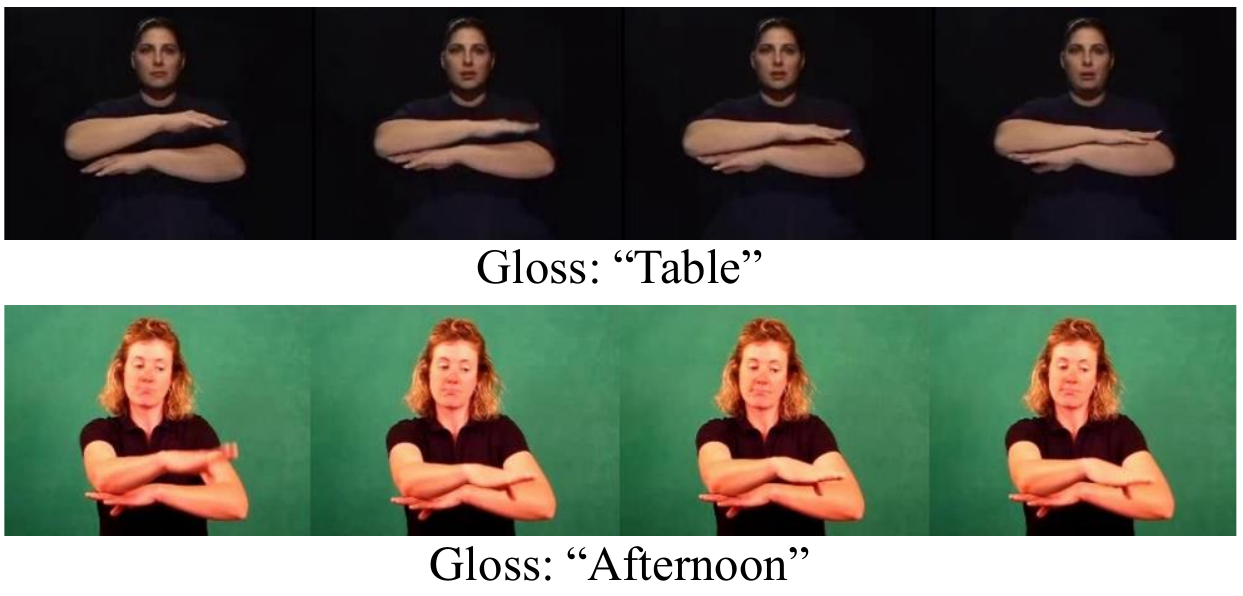}
         \caption{VISigns may have \textit{distinct} semantic meanings.}
         \label{fig:teaser_B}
     \end{subfigure}
     \vspace{-2mm}
    \caption{Vision neural networks are demonstrated to be less effective to recognize visually indistinguishable signs (VISigns)~\cite{albanie2020bsl, li2020word, joze2019ms}. We observe that VISigns may have similar or distinct semantic meanings, inspiring us to leverage this characteristic to facilitate sign language recognition as illustrated in Figure~\ref{fig:teaser_vl}.}
    \vspace{-6mm}
    \label{fig:teaser}
\end{figure}

\begin{figure*}
     \centering
     \begin{subfigure}[b]{0.43\textwidth}
         \centering
         \includegraphics[width=\textwidth]{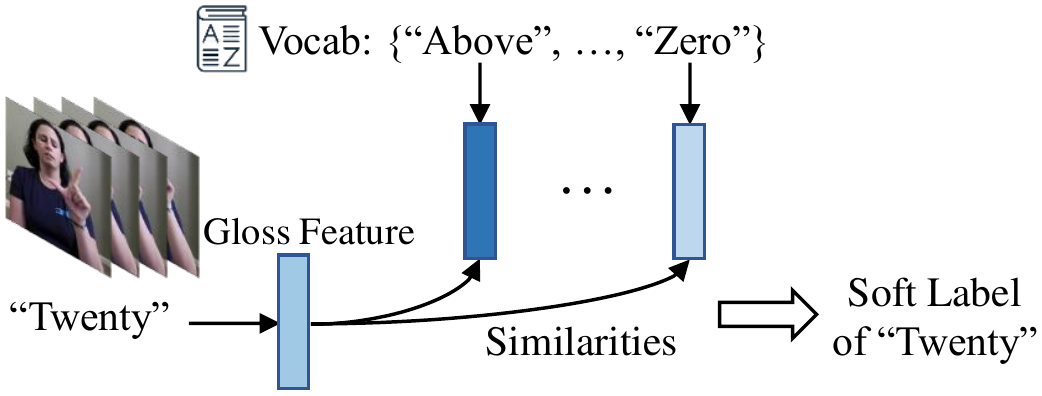}
         \caption{Language-aware label smoothing.}
         \label{fig:teaser_D}
     \end{subfigure}
     \hfill
     \begin{subfigure}[b]{0.51\textwidth}
         \centering
         \includegraphics[width=\textwidth]{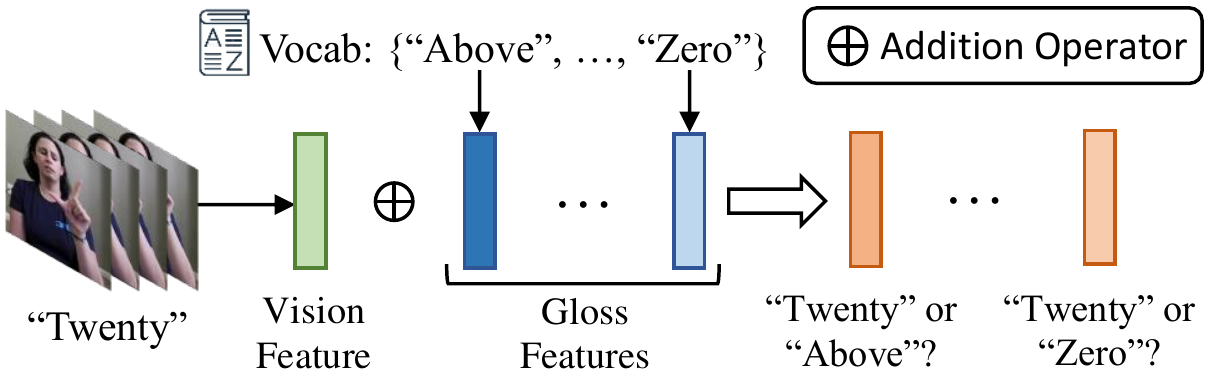}
         \caption{Inter-modality mixup.}
         \label{fig:teaser_E}
     \end{subfigure}
     \vspace{-3mm}
    \caption{We incorporate natural language modeling into sign language recognition to promote recognition capacity. (a) Language-aware label smoothing generates a soft label for each training video, whose smoothing weights are the normalized semantic similarities of the ground truth gloss and the remaining glosses within the sign language vocabulary. (b) Inter-modality mixup yields the blended features (denoted by orange rectangles) with the corresponding mixed labels to maximize the separability of signs in a latent space.}
    \vspace{-5mm}
    \label{fig:teaser_vl}
\end{figure*}

Since the lexical items of sign languages are defined by the handshape, facial expression, and movement, the combinations of these visual ingredients are restricted inherently, yielding plenty of visually indistinguishable signs termed VISigns. VISigns are those signs with similar handshape and motion but varied semantic meanings. We show two examples (``Cold'' \vs ``Winter'' and ``Table'' \vs ``Afternoon'') in Figure~\ref{fig:teaser}. Unfortunately, vision neural networks are demonstrated to be less effective to accurately recognize the VISigns~\cite{albanie2020bsl, li2020word, joze2019ms}. Due to the intrinsic connections between sign languages and natural languages, the glosses, \ie, labels of signs, are semantically meaningful in contrast to the one-hot labels used in traditional classification tasks~\cite{kay2017kinetics, imagenet}. 
Thus, although the VISigns are challenging to be classified from the vision perspective, their glosses provide
serviceable semantics, which is, however, less taken into consideration in previous works~\cite{hu2021hand, hu2021signbert, hu2021global, li2020transferring, li2020word, joze2019ms, jiang2021skeleton, jiang2021sign}. Our work is built upon the following two findings.

\textit{Finding-1: VISigns may have similar semantic meanings (Figure~\ref{fig:teaser_A}).}
Due to the observation that VISigns may have higher visual similarities, assigning hard labels to them may hinder the training since it is challenging for vision neural networks to distinguish each VISign apart. 
A straightforward way to ease the training is to replace the hard labels with soft ones as in well-established label smoothing~\cite{szegedy2016rethinking, he2019bag}.
However, how to generate proper soft labels is non-trivial.
The vanilla label smoothing~\cite{szegedy2016rethinking, he2019bag} assigns equal smoothing weights to all negative terms, which ignores the semantic information contained in labels.
In light of the \textit{finding-1} that VISigns may have similar semantic meanings and the intrinsic connections between sign languages and natural languages, we consider the semantic similarities among the glosses when generating soft labels. 
Concretely, for each training video, we adopt an off-the-shelf word representation framework, \ie, fastText~\cite{mikolov2018advances}, to pre-compute the semantic similarities of its gloss and the remaining glosses within the sign language vocabulary. 
Then we can properly generate a soft label for each training sample whose smoothing weights are the normalized semantic similarities. 
In this way, negative terms with similar semantic meanings to the ground truth gloss are assigned higher values in the soft label.
As shown in Figure~\ref{fig:teaser_D}, we term this process as language-aware label smoothing, which injects prior knowledge into the training.

\textit{Finding-2: VISigns may have distinct semantic meanings (Figure~\ref{fig:teaser_B}).} Although the VISigns are challenging to be classified from the vision perspective, the semantic meanings of their glosses may be distinguishable according to \textit{finding-2}. This inspires us to combine the vision features and gloss features to drive the model towards maximizing signs' separability in a latent space. Specifically, given a sign video, we first leverage our proposed backbone to encode its vision feature and the well-established fastText~\cite{mikolov2018advances} to extract the feature of each gloss within the sign language vocabulary. Then we independently integrate the vision feature and each gloss feature to produce a blended representation, which is further fed into a classifier to approximate its mixed label. We refer to this procedure as inter-modality mixup as shown in Figure~\ref{fig:teaser_E}. We empirically find that our inter-modality mixup significantly enhances the model's discriminative power.

Our contributions can be summarized as follows:
\vspace{-2mm}
\begin{itemize}
\setlength{\itemsep}{0pt}
\setlength{\parsep}{0pt}
\setlength{\parskip}{3pt}
    \item We are the first to incorporate natural language modeling into sign language recognition based on the discovery of VISigns. Language-aware label smoothing and inter-modality mixup are proposed to take full advantage of the linguistic properties of VISigns and semantic information contained in glosses.
    \item We take into account the unique characteristic of sign languages and present a novel backbone named video-keypoint network (VKNet), which not only models both RGB videos and human keypoints, but also derives knowledge from sign videos of various temporal receptive fields.
    \item Our method, termed natural language-assisted sign language recognition (NLA-SLR), achieves state-of-the-art performance on the widely-used SLR datasets including MSASL~\cite{joze2019ms}, WLASL~\cite{li2020word}, and NMFs-CSL~\cite{hu2021global}.
\end{itemize}

\begin{figure*}[t]
\centering
\includegraphics[width=0.98\linewidth]{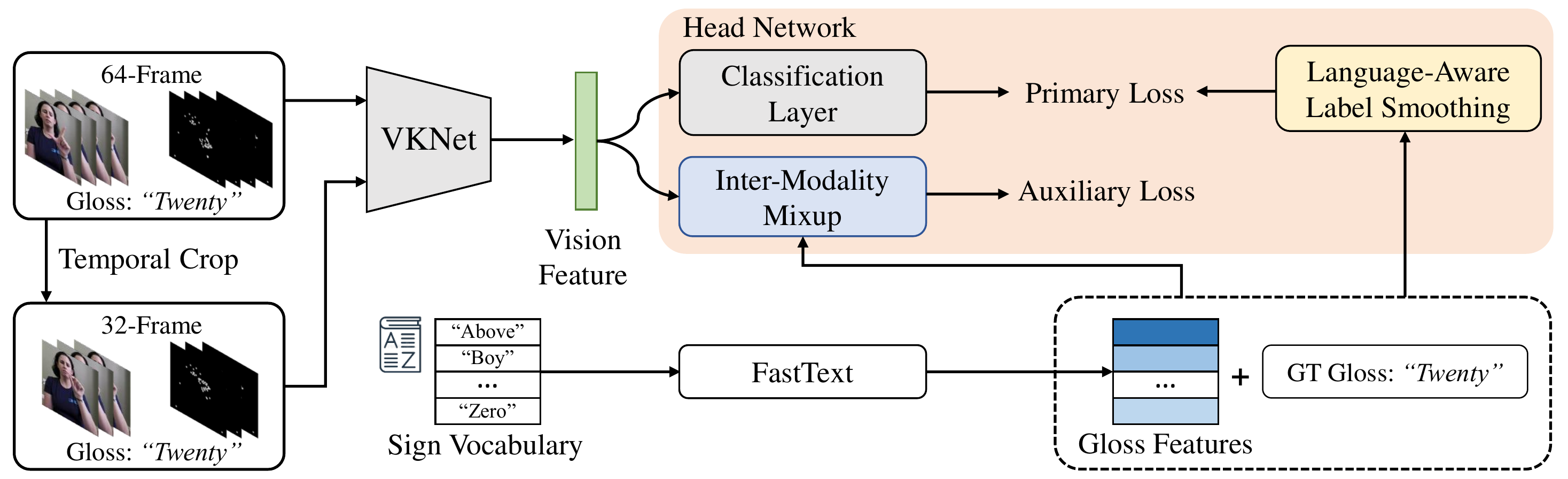}
\vspace{-3mm}
\caption{An overview of our NLA-SLR. Given a training video, we temporally crop a 64-frame clip \cite{li2020transferring} and use HRNet~\cite{sun2019deep} trained on COCO-WholeBody~\cite{jin2020whole} to estimate its keypoint sequence which is represented by a set of heatmaps, yielding a 64-frame video-keypoint pair. Then we temporally crop a 32-frame counterpart and feed it along with the 64-frame pair into our proposed VKNet (Figure~\ref{fig:vknet}) to extract the vision feature. The head network (Figure~\ref{fig:vl_mixup}) has a two-branch architecture consisting of a language-aware label smoothing branch and an inter-modality mixup branch. We only retain the VKNet and the classification layer in the head network for inference.}
\vspace{-5mm}
\label{fig:framework}
\end{figure*}

\vspace{-4mm}
\section{Related Works}
\noindent\textbf{Sign Language Recognition.} Sign language recognition (SLR) is a fundamental task in the field of sign language understanding.
Feature extraction plays a key role in an SLR model.
Most recent SLR works \cite{jiang2021sign, jiang2021skeleton, hu2021signbert, hu2021hand, li2020transferring, li2020word, joze2019ms, hu2021global, stmc, zuo22_interspeech, vac} adopt CNN-based architectures, \eg, I3D \cite{I3D} and R3D \cite{qiu2017learning}, to extract vision features from RGB videos.
In this work, we adopt S3D \cite{xie2018rethinking} as the backbone of our VKNet due to its excellent accuracy-speed trade-off.

However, RGB-based SLR models may suffer from the large variation of video backgrounds. 
As a complement, some SLR works \cite{jiang2021skeleton, jiang2021sign, hu2021hand, hu2021signbert, chentwo} explore to jointly model RGB videos and keypoints.
For example, SAM-SLR \cite{jiang2021skeleton} uses graph convolutional networks (GCNs) to model pre-extracted keypoints.
HMA \cite{hu2021hand} and SignBERT \cite{hu2021signbert} propose to decode 3D hand keypoints from RGB videos.
A common deficiency of these works is that they need a dedicated network to model keypoints.
In this work, we represent keypoints as a sequence of heatmaps~\cite{duan2022revisiting, chentwo} so that the keypoint encoder of our VKNet can share the identical architecture with the video encoder.

To enable mini-batch training, previous works \cite{jiang2021sign, jiang2021skeleton, hu2021signbert, hu2021hand, li2020transferring, li2020word} crop fixed-length clips from raw videos as model inputs.
However, the model may overfit to the training videos of fixed temporal receptive fields.
In contrast, our VKNet is trained on videos with varied temporal receptive fields to improve its generalization capability.

\noindent\textbf{Word Representation Learning.}
Word2vec \cite{word2vec} and GloVe \cite{glove} are two classical word representation learning frameworks in the field of NLP.
Based on word2vec, fastText \cite{mikolov2018advances} improves word representations with several modifications including the use of sub-word information \cite{bojanowski2017enriching} and position independent features \cite{mnih2013learning}.
Although some advanced language models, \eg, BERT \cite{kenton2019bert}, can also be used to extract word representations, they are computationally intensive and are not dedicated to word representation learning.
In this paper, we adopt the lightweight but effective fastText, which is also used in a recent sign language translation work \cite{yin2021simulslt}, to pre-compute gloss (word) representations.

\noindent\textbf{Vision-Language Models.}
Recently, a majority of vision-language models \cite{clip, align, yao2022filip, gu2022wukong} learn visual representations on large-scale image-text pairs.
Among them, CLIP \cite{clip} is the pioneer to jointly optimize an image encoder and a text encoder through a contrastive loss. 
Besides, the pre-trained CLIP can be generalized to various downstream tasks, \eg, semantic segmentation \cite{xu2022groupvit, li2021language, xu2021simple}, object detection \cite{du2022learning, rao2022denseclip}, image classification~\cite{zhou2022learning,huang2022unsupervised}, and style transfer \cite{patashnik2021styleclip, kwon2022clipstyler}.
In this work, we exploit the implicit knowledge included in glosses (sign labels), which is distinct from previous works on vision-language modeling.

\noindent\textbf{Multi-label Classification.} Real-world objects may have multiple semantic meanings, which motivates research on multi-label classification \cite{ridnik2021asymmetric, ke2022hyperspherical, zhang2013review, rajeswar2022multi, kim2022large} requiring models to map inputs to multiple possible labels.
Although the VISigns may be associated with the multi-label classification problem, most widely-adopted SLR datasets \cite{li2020word, joze2019ms, hu2021global} are singly labeled.
In this work, we deal with the VISigns by incorporating language information included in glosses.
\section{Methodology}
An overview of our natural language-assisted sign language recognition (NLA-SLR) framework is shown in Figure \ref{fig:framework}. Our framework mainly consists of three parts: 1) data pre-processing which generates video-keypoint pairs as network inputs (Section~\ref{sec:data}); 2) a video-keypoint network (VKNet) which takes video-keypoint pairs of various temporal receptive fields as inputs for vision feature extraction (Section~\ref{sec:vknet}); 3) a head network (Section~\ref{sec:head}) containing a language-aware label smoothing branch (Section~\ref{sec:lang_lbsm}) and an inter-modality mixup branch (Section~\ref{sec:vl_mixup}). We empirically find that Mixup~\cite{zhang2018mixup} can be applied on both RGB videos and keypoint heatmap sequences, which will be described in Section~\ref{sec:intra}.

\subsection{Data Pre-Processing}
\label{sec:data}
Sign languages are visual languages which adopt handshape, facial expression, and body movement to convey information.
To more effectively model sign languages, we propose to model human body keypoints besides RGB videos to enhance the robustness of visual representations. 

Concretely, given a temporally cropped video $\boldsymbol{V} \in \mbb{R}^{T\times H_{V} \times W_{V} \times 3}$ with $T=64$ frames \cite{li2020transferring} and a spatial resolution of $H_V=W_V=224$, we use HRNet~\cite{sun2019deep} trained on COCO-WholeBody~\cite{jin2020whole} to estimate its 63 keypoints (11 for upper body, 10 for mouth, and 42 for two hands) per frame. The keypoints of the $t$-th frame are represented as a heatmap $\boldsymbol{K}_t\in\mathbb{R}^{H_K \times W_K \times K}$, where $H_K=W_K=112$ denote the height and width of the heatmap, and $K=63$ is the keypoint number. The elements within the heatmap $\boldsymbol{K}_t$ are generated by a Gaussian function: $\boldsymbol{K}_t[i,j,k] = \exp (-[(i-x_t^k)^2 + (j-y_t^k)^2] / 2\sigma^2)$, where $(i, j)$ represents the spatial index, $k$ is the keypoint index, $(x_t^k, y_t^k)$ denotes the coordinate of the $k$-th estimated keypoint of the $t$-th frame, and $\sigma=4$ controls the scale of the keypoints. We repeatedly generate the heatmaps for all frames and stack them along the temporal dimension into a keypoint heatmap sequence $\boldsymbol{K} \in \mathbb{R}^{T \times H_K \times W_K \times K}$. Now the 64-frame training sample is processed as a video-keypoint pair denoted as $(\boldsymbol{V}_{64}, \boldsymbol{K}_{64})$. Finally, we temporally crop a 32-frame counterpart $(\boldsymbol{V}_{32}, \boldsymbol{K}_{32})$ and feed it along with the 64-frame video-keypoint pair $(\boldsymbol{V}_{64}, \boldsymbol{K}_{64})$ into the VKNet to extract more robust vision features, which will be described in the next section.

\subsection{Video-Keypoint Network}
\label{sec:vknet}
\begin{figure}[t]
\centering
\includegraphics[width=1.0\linewidth]{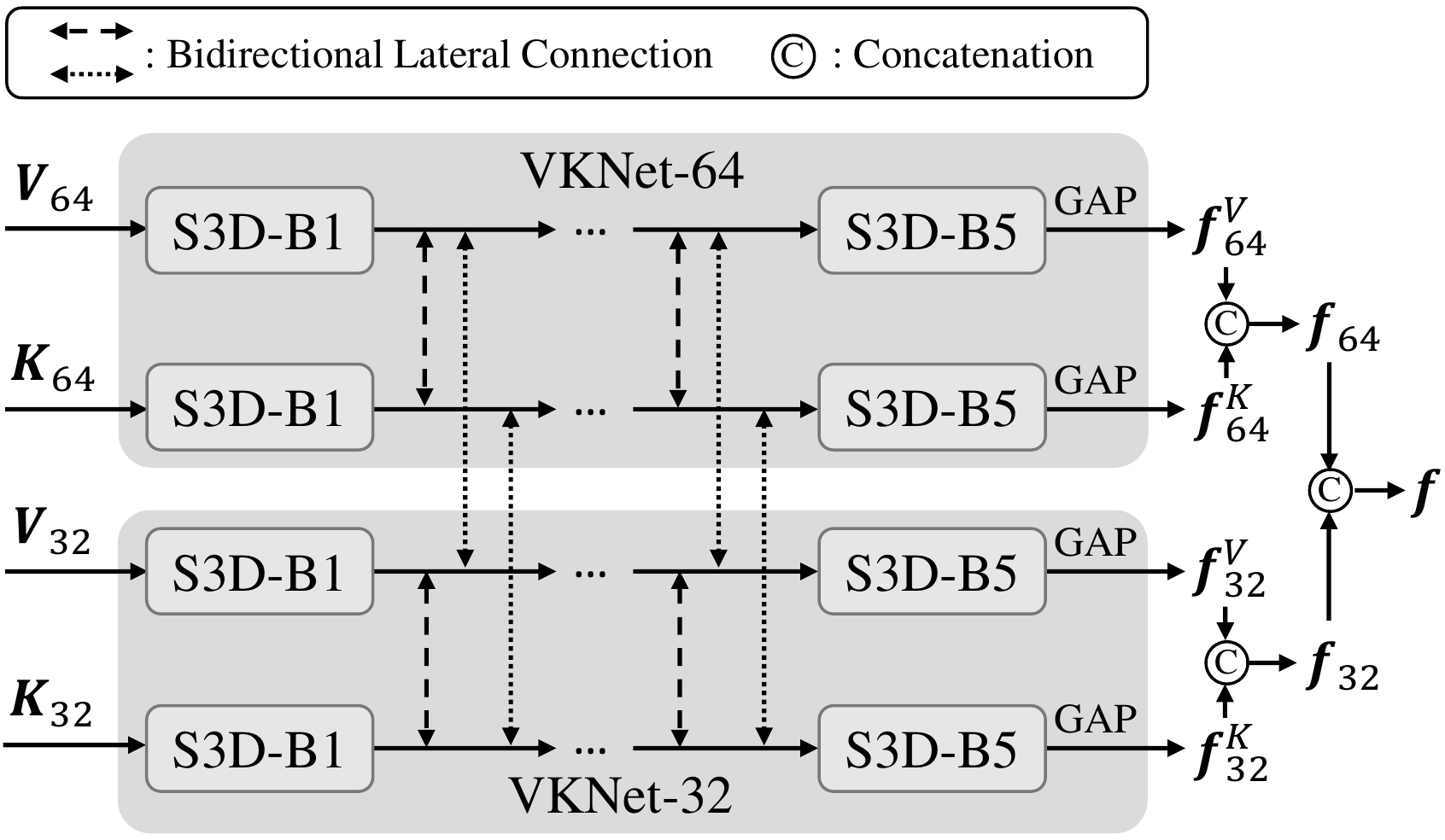}
\vspace{-7mm}
\caption{Our VKNet consists of two sub-networks, VKNet-64 and VKNet-32, which take video-keypoint pairs with different temporal receptive fields as inputs and output a set of vision features via global average pooling (GAP) layers. Within the VKNet, bidirectional lateral connections \cite{duan2022revisiting} are applied to the outputs of the first four S3D blocks (B1-B4) for video-video, keypoint-keypoint, and video-keypoint information exchange.}
\vspace{-5mm}
\label{fig:vknet}
\end{figure}

An illustration of the proposed video-keypoint network (VKNet) is shown in Figure~\ref{fig:vknet}. VKNet is composed of two sub-networks, namely VKNet-32 and VKNet-64, which take $(\boldsymbol{V}_{32}, \boldsymbol{K}_{32})$ and $(\boldsymbol{V}_{64}, \boldsymbol{K}_{64})$ as inputs, respectively. The network architectures of VKNet-32 and VKNet-64 are identical—either has a two-stream architecture consisting of a video encoder and a keypoint encoder. Since we denote keypoints as heatmaps, it is feasible to utilize any existing convolutional neural networks to extract keypoint features. In this work, S3D \cite{xie2018rethinking} with five blocks (B1--B5) is served as our video/keypoint encoder due to its excellent accuracy-speed trade-off. In our implementation, VKNet-32 (VKNet-64) is composed of two separate S3D networks with bidirectional lateral connections \cite{duan2022revisiting} applied to the outputs of the first four blocks (B1--B4). Specifically, VKNet-32 (VKNet-64) takes RGB video $\boldsymbol{V}_{32}$ ($\boldsymbol{V}_{64}$) and keypoint heatmap sequence $\boldsymbol{K}_{32}$ ($\boldsymbol{K}_{64}$) as inputs to extract the video feature $\boldsymbol{f}_{32}^{V}$ ($\boldsymbol{f}_{64}^{V}$) and the keypoint feature $\boldsymbol{f}_{32}^{K}$ ($\boldsymbol{f}_{64}^{K}$), respectively. 
We further concatenate $\boldsymbol{f}_{32}^{V}$ ($\boldsymbol{f}_{64}^{V}$) and $\boldsymbol{f}_{32}^{K}$ ($\boldsymbol{f}_{64}^{K}$) to generate $\boldsymbol{f}_{32}$ ($\boldsymbol{f}_{64}$) as the output of VKNet-32 (VKNet-64). The final feature $\boldsymbol{f}$ extracted by VKNet is the concatenation of $\boldsymbol{f}_{32}$ and $\boldsymbol{f}_{64}$.

It is worth mentioning that VKNet-32 and VKNet-64 are not two independent networks, we also introduce bidirectional lateral connections \cite{duan2022revisiting} to the corresponding encoders of the same input modality for video-video and keypoint-keypoint information exchange.

\begin{figure}[t]
\centering
\includegraphics[width=1.0\linewidth]{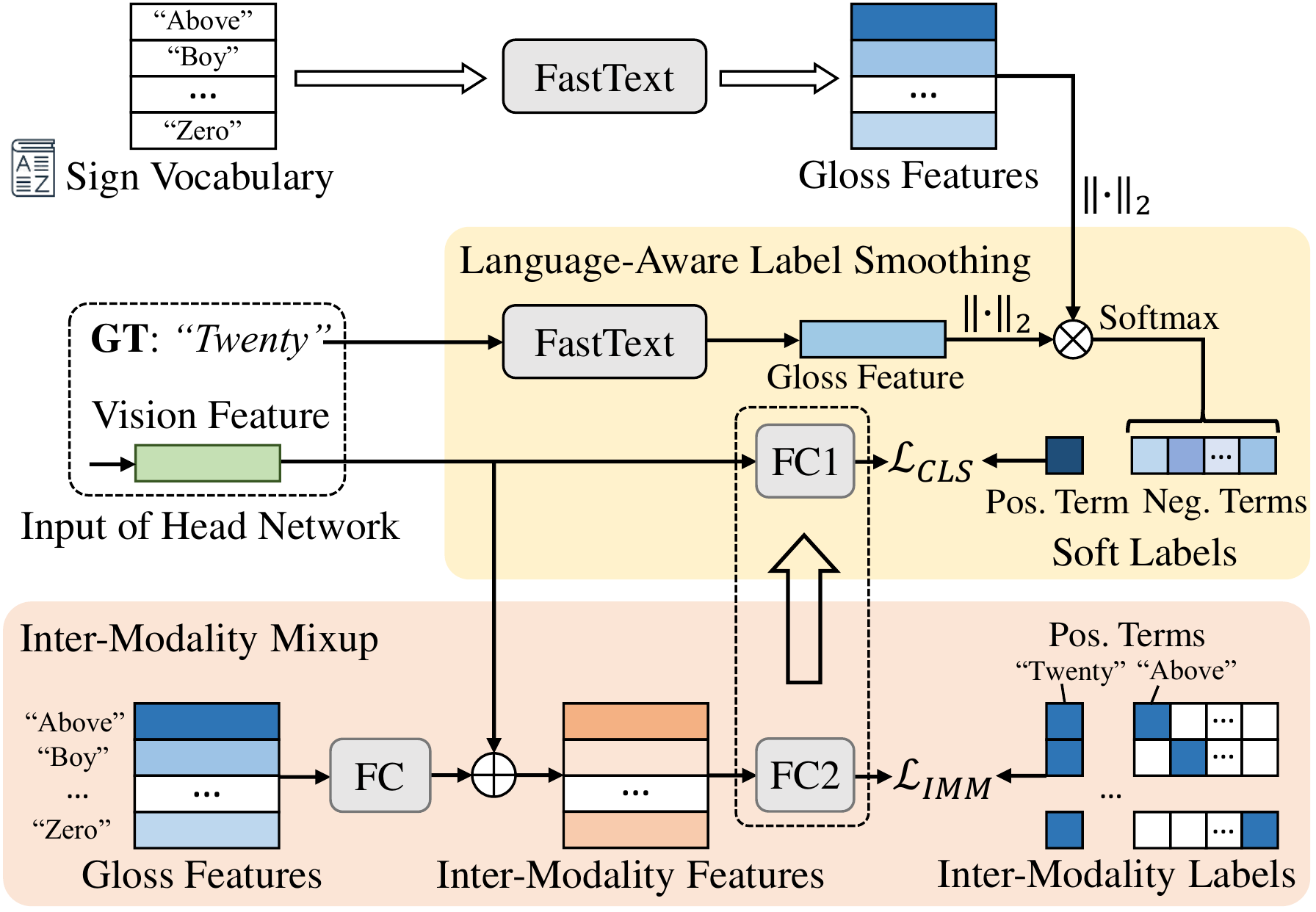}
\vspace{-7mm}
\caption{
The architecture of our head network. Language-aware label smoothing generates soft labels whose smoothing weights are the normalized semantic similarities between the ground truth and remaining glosses within the sign vocabulary. Inter-modality mixup generates inter-modality features and the corresponding labels to maximize the signs' separability in a latent space.
Integration between FC1 and FC2 can further boost SLR performance.
}
\vspace{-5mm}
\label{fig:vl_mixup}
\end{figure}

\subsection{Head Network}
\label{sec:head}
Figure~\ref{fig:vl_mixup} illustrates our head network, which is composed of a language-aware label smoothing branch and an inter-modality mixup branch.

\vspace{-4mm}
\subsubsection{Language-Aware Label Smoothing}
\vspace{-2mm}
\label{sec:lang_lbsm}
The classical label smoothing \cite{szegedy2016rethinking, he2019bag} was first proposed as a regularization technique to alleviate overfitting and make the model more adaptable.
Specifically, given a training sample belonging to the $b$-th class, label smoothing replaces the one-hot label with a soft label $\boldsymbol{y} \in \mathbb{R}^{N}$ which is defined as:
\begin{equation}
\label{equ:vani_lbsm}
\boldsymbol{y}[i]=
\begin{cases}
    1-\epsilon \ &\text{if}\  i=b, \\
    \epsilon/(N-1) \  &\text{otherwise},
\end{cases}
\end{equation}
where $\epsilon$ is a small constant (\eg, 0.2) and $N$ denotes the class number.

The vanilla label smoothing uniformly distributes $\epsilon$ to $N-1$ negative terms while the implicit semantics contained in glosses (sign labels) are ignored. In Section~\ref{sec:intro}, we discuss the phenomenon that visually indistinguishable signs (VISigns) may have similar semantic meanings (\textit{finding-1}). Motivated by this finding, we present a novel regularization strategy termed language-aware label smoothing, which assigns biased smoothing weights on the basis of semantic similarities of glosses to ease the training.

\noindent\textbf{Gloss Features.} Gloss is identified by a word which is associated with the sign’s semantic meaning. Thus any word representation learning framework can be adopted to extract gloss features for semantic similarity assessment. Concretely, given a sign vocabulary containing $N$ glosses, we leverage fastText~\cite{mikolov2018advances} pretrained on Common Crawl to extract a 300-$d$ feature for each gloss. We use $\boldsymbol{E}\in\mathbb{R}^{N \times 300}$ to denote the $N$ gloss features.

\noindent\textbf{Language-Aware Label Smoothing and Loss Function.} As shown in Figure~\ref{fig:vl_mixup}, given a training sample whose label is the $b$-th gloss, we first use fastText to extract its gloss feature $\boldsymbol{e} \in \mathbb{R}^{300}$. Then we compute the cosine similarities of the $b$-th gloss and all $N$ glosses within the sign vocabulary by $\boldsymbol{s}=\|\boldsymbol{E}\|_2 \|\boldsymbol{e}\|_2^T \in \mathbb{R}^{N}$, where $\|\cdot\|_2$ denotes row-wise L2-norm. The proposed language-aware label smoothing generates a soft label $\boldsymbol{y} \in \mathbb{R}^{N}$ as:
\begin{equation}
\label{equ:lang_lbsm}
\boldsymbol{y}[i]=
\begin{cases}
    1-\epsilon \ &\text{if}\  i=b, \\
    \epsilon \cdot \frac{\exp{(\boldsymbol{s}[i]/\tau)}}{\sum_{i=1,i\neq b}^N \exp{(\boldsymbol{s}[i]/\tau)}} \  &\text{otherwise},
\end{cases}
\end{equation}
where $\tau$ denotes a temperature parameter \cite{chen2020simple}.
The classification loss $\mathcal{L}_{CLS}$ is a simple cross-entropy loss applied on the prediction and soft label $\boldsymbol{y}$.

\vspace{-4mm}
\subsubsection{Inter-Modality Mixup}
\vspace{-2mm}
\label{sec:vl_mixup}
In Section~\ref{sec:intro}, we observe that VISigns may have distinct semantic meanings (\textit{finding-2}), motivating us to make use of the semantic meanings of glosses to maximize signs' separability in the latent space. To achieve the goal, as shown in Figure~\ref{fig:vl_mixup}, we introduce the inter-modality mixup, which generates the inter-modality features by combining the vision feature and gloss features to predict the corresponding inter-modality labels.

\noindent\textbf{Inter-Modality Mixup and Loss Function.} Given the vision feature $\boldsymbol{f}\in \mathbb{R}^{D}$ extracted by our VKNet and the gloss features $\boldsymbol{E} \in \mathbb{R}^{N\times 300}$ encoded by the fastText, we first use a fully-connected (FC) layer to map $\boldsymbol{E}$ to the dimension of $N \times D$. After that, we integrate the vision feature $\boldsymbol{f}$ and the mapped gloss features $\bar{\boldsymbol{E}}$ via a broadcast addition operation into the inter-modality features $\boldsymbol{F} = \boldsymbol{f} + \bar{\boldsymbol{E}} \in\mathbb{R}^{N \times D}$. The $n$-th row of $\boldsymbol{F}$ (denoted as $\boldsymbol{F}^n$), which is the combination of the vision feature (whose corresponding ground truth is the $b$-th gloss) and the $n$-th gloss feature, is associated with the inter-modality labels $\boldsymbol{y}^n \in \mathbb{R}^{N}$:
\begin{equation}
\boldsymbol{y}^n[i] = 
\begin{cases}
    0.5 \quad \text{if}\  i=b~\text{or}~i=n, \\
    0 ~~~\quad \text{otherwise}.
\end{cases}
\end{equation}
Note that as a special case, we set $\boldsymbol{y}^n[b]=1.0$ when $n=b$. Then we feed $\boldsymbol{F}^n$ into a classification layer to generate its prediction $\boldsymbol{p}^n \in (0,1)^{N}$, and use cross-entropy loss to approximate $\boldsymbol{y}^n$:
\begin{equation}
    \mathcal{L}_{IMM}^{n} = -\sum_{i=1}^{N}\boldsymbol{y}^n[i]\text{log}(\boldsymbol{p}^n[i]).
\end{equation}
Similarly, we could obtain the predictions of $N$ inter-modality features and their corresponding labels. The overall loss of inter-modality mixup is the average of $N$ cross-entropy losses:
\begin{equation}
    \mathcal{L}_{IMM} = \frac{1}{N}\sum_{n=1}^{N}\mathcal{L}_{IMM}^{n}.
\end{equation}
It is worth noting that $\mathcal{L}_{IMM}$ is an auxiliary loss and we drop the inter-modality mixup branch in the inference stage.

\noindent\textbf{Boost Sign Language Recognition via the Integrated Classification Layer.} As shown in Figure \ref{fig:vl_mixup}, we term the classification layer in the language-aware label smoothing branch and inter-modality mixup branch as FC1 and FC2, respectively. Though the inter-modality mixup only attends the training, the well-optimized FC2 contains implicit knowledge of recognizing signs with the help of language information. This inspires us to integrate FC2 into FC1 to boost sign language recognition. Concretely, the parameters of the FC1 are updated by a weighted sum of its own parameters and the FC2's parameters at each iteration, which can be formulated as:
\begin{equation}
\label{equ:ema}
\begin{split}
\theta_1, \theta_2 &\leftarrow \text{optimizer}(\theta_1, \theta_2, \nabla_{\theta_1}\mathcal{L}, \nabla_{\theta_2}\mathcal{L}, \eta) \\
\theta_1 &\leftarrow \mu\theta_1 + (1-\mu)\theta_2,
\end{split}
\end{equation}
where $\theta_1$ and $\theta_2$ denote the parameters of FC1 and FC2, respectively, $\mathcal{L}$ is the overall loss of the head network introduced in Section \ref{sec:overall_loss}, $\eta$ is the learning rate, and $\mu$ controls the contribution of $\theta_2$.

\begin{table*}[t]
\setlength\tabcolsep{4pt}
\centering
\resizebox{\linewidth}{!}{
\begin{tabular}{l|cc|cc|cc|cc|cc|cc|cc|cc}
\toprule
\multirow{3}{*}{Method} & \multicolumn{4}{c|}{MSASL1000} & \multicolumn{4}{c|}{MSASL500} & \multicolumn{4}{c|}{MSASL200} & \multicolumn{4}{c}{MSASL100} \\
\cmidrule(){2-17}
& \multicolumn{2}{c|}{Per-instance} & \multicolumn{2}{c|}{Per-class} & \multicolumn{2}{c|}{Per-instance} & \multicolumn{2}{c|}{Per-class} & \multicolumn{2}{c|}{Per-instance} & \multicolumn{2}{c|}{Per-class} & \multicolumn{2}{c|}{Per-instance} & \multicolumn{2}{c}{Per-class} \\
& Top-1 & Top-5 & Top-1 & Top-5 & Top-1 & Top-5 & Top-1 & Top-5 & Top-1 & Top-5 & Top-1 & Top-5 & Top-1 & Top-5 & Top-1 & Top-5 \\

\midrule
I3D \cite{I3D} & -- & -- & 57.69 & 81.08 & -- & -- & 72.50 & 89.80 & -- & -- & 81.97 & 93.79 & -- & -- & 81.76 & 95.16 \\
I3D+BLSTM \cite{I3D,lstm} & 40.99 & -- & -- & -- & -- & -- & -- & -- & -- & -- & -- & -- & 72.07 & -- & -- & -- \\
ST-GCN \cite{yan2018spatial} & 36.03 & 59.92 & 32.32 & 57.15 & -- & -- & -- & -- & 52.91 & 76.67 & 54.20 & 77.62 & 59.84 & 82.03 & 60.79 & 82.96 \\
BSL (multi-crop) \cite{albanie2020bsl} & 64.71 & 85.59 & 61.55 & 84.43 & -- & -- & -- & --  & -- & --  & -- & --  & -- & --  & -- & -- \\
TCK$\dagger$ \cite{li2020transferring} & -- & -- & -- & -- & -- & -- & -- & -- & 80.31 & 91.82 & 81.14 & 92.24 & 83.04 & 93.46 & 83.91 & 93.52 \\
HMA \cite{hu2021hand} & 69.39 & 87.42 & 66.54 & 86.56 & -- & -- & -- & -- & 85.21 & 94.41 & 86.09 & 94.42  & 87.45 & 96.30  & 88.14 & 96.53 \\
BEST \cite{best} & 71.21 & 88.85 & 68.24 & 87.98 & -- & -- & -- & -- & 86.83 & 95.66 & 87.45 & 95.72 & 89.56 & 96.96 & 90.08 & 97.07 \\
SignBERT$\dagger$ \cite{hu2021signbert} & 71.24 & 89.12 & 67.96 & 88.40 & -- & -- & -- & -- & 86.98 & 96.39 & 87.62 & 96.43 & 89.56 & 97.36 & 89.96 & 97.51 \\
\midrule

NLA-SLR (Ours) & 72.56 & 89.12 & 69.86 & 88.48 & 81.62 & 93.09 & 81.36 & 93.39 & 88.74 & 96.17 & 89.23 & 96.38 & 90.49 & 97.49 & 91.04 & 97.92 \\
NLA-SLR (Ours, 3-crop) & \tbf{73.80} & \tbf{89.65} & \tbf{70.95} & \tbf{89.07} & \tbf{82.90} & \tbf{93.46} & \tbf{83.06} & \tbf{93.54} & \tbf{89.48} & \tbf{96.69} & \tbf{89.86} & \tbf{96.93} & \tbf{91.02} & \tbf{97.89} & \tbf{91.24} & \tbf{98.19} \\
\bottomrule
\end{tabular}}
\vspace{-3mm}
\caption{Comparison with previous works on MSASL. The results of I3D, I3D+BLSTM, and ST-GCN are reproduced by \cite{joze2019ms}, \cite{adaloglou2021comprehensive}, and \cite{hu2021signbert}, respectively. BSL achieves multi-crop inference by sliding a window with a stride of 8 frames. ($\dagger$denotes methods using extra data.)}
\vspace{-5mm}
\label{tab:sota_msasl}
\end{table*}

\vspace{-4mm}
\subsubsection{Overall Loss}
\vspace{-2mm}
\label{sec:overall_loss}
The loss $\mathcal{L}$ of the head network is the sum of the classification loss $\mathcal{L}_{CLS}$ and the inter-modality mixup loss $\mathcal{L}_{IMM}$ with a trade-off hyper-parameter $\gamma$: $\mathcal{L} = \mathcal{L}_{CLS} + \gamma \mathcal{L}_{IMM}$.
Note that we apply the head network to each vision feature in Figure \ref{fig:vknet} independently, and the overall loss for the whole model is the sum of the loss of each head network.

\subsection{Intra-Modality Mixup}
\label{sec:intra}
We empirically find that Mixup~\cite{zhang2018mixup} is helpful for sign language recognition. In contrast to the traditional Mixup which is applied to images and videos, we adopt the Mixup regularization on both RGB videos and keypoint heatmap sequences. For a distinction with our proposed Inter-Modality Mixup, we term the classical Mixup as Intra-Modality Mixup in our work.

\section{Experiments}
\begin{table*}[t]
\setlength\tabcolsep{4pt}
\centering
\resizebox{\linewidth}{!}{
\begin{tabular}{l|cc|cc|cc|cc|cc|cc|cc|cc}
\toprule
\multirow{3}{*}{Method} & \multicolumn{4}{c|}{WLASL2000} & \multicolumn{4}{c|}{WLASL1000} & \multicolumn{4}{c|}{WLASL300} & \multicolumn{4}{c}{WLASL100} \\
\cmidrule(){2-17}
& \multicolumn{2}{c|}{Per-instance} & \multicolumn{2}{c|}{Per-class} & \multicolumn{2}{c|}{Per-instance} & \multicolumn{2}{c|}{Per-class} & \multicolumn{2}{c|}{Per-instance} & \multicolumn{2}{c|}{Per-class} & \multicolumn{2}{c|}{Per-instance} & \multicolumn{2}{c}{Per-class} \\
& Top-1 & Top-5 & Top-1 & Top-5 & Top-1 & Top-5 & Top-1 & Top-5 & Top-1 & Top-5 & Top-1 & Top-5 & Top-1 & Top-5 & Top-1 & Top-5 \\

\midrule
OpenHands \cite{selvaraj2022openhands} & 30.60 & -- & -- & -- & -- & -- & -- & -- & -- & -- & -- & -- & -- & -- & -- & -- \\
PSLR \cite{tunga2021pose} & -- & -- & -- & -- & -- & -- & -- & -- & 42.18 & 71.71 & -- & -- & 60.15 & 83.98 & -- & -- \\
I3D \cite{I3D} & 32.48 & 57.31 & -- & -- & 47.33 & 76.44 & -- & -- & 56.14 & 79.94 & -- & -- & 65.89 & 84.11 & -- & -- \\
ST-GCN \cite{yan2018spatial} & 34.40 & 66.57 & 32.53 & 65.45 & -- & -- & -- & -- & 44.46 & 73.05 & 45.29 & 73.16 & 50.78 & 79.07 & 51.62 & 79.47 \\
Fusion-3 \cite{hosain2021hand} & 38.84 & 67.58 & -- & -- & 56.68 & 79.85 & -- & -- & 68.30 & 83.19 & -- & -- & 75.67 & 86.00 & -- & -- \\
BSL (multi-crop) \cite{albanie2020bsl} & 46.82 & 79.36 & 44.72 & 78.47 & -- & -- & -- & --  & -- & --  & -- & --  & -- & --  & -- & -- \\
HMA \cite{hu2021hand} & 51.39 & 86.34 & 48.75 & 85.74 & -- & -- & -- & --  & -- & --  & -- & --  & -- & --  & -- & -- \\
TCK$\dagger$ \cite{li2020transferring} & -- & -- & -- & -- & -- & -- & -- & -- & 68.56 & 89.52 & 68.75 & 89.41 & 77.52 & 91.08 & 77.55 & 91.42 \\
BEST \cite{best} & 54.59 & 88.08 & 52.12 & 87.28 & -- & -- & -- & -- & 75.60 & 92.81 & 76.12 & 93.07 & 81.01 & 94.19 & 81.63 & 94.67 \\
SignBERT$\dagger$ \cite{hu2021signbert} & 54.69 & 87.49 & 52.08 & 86.93 & -- & -- & -- & -- & 74.40 & 91.32 & 75.27 & 91.72 & 82.56 & 94.96 & 83.30 & 95.00 \\
SAM-SLR* (5-crop) \cite{jiang2021skeleton} & 58.73 & 91.46 & 55.93 & \tbf{90.94} & -- & -- & -- & -- & -- & -- & -- & -- & -- & -- & -- & -- \\
SAM-SLR-v2* (5-crop) \cite{jiang2021sign} & 59.39 & 91.48 & 56.63 & 90.89 & -- & -- & -- & -- & -- & -- & -- & -- & -- & -- & -- & -- \\
\midrule

NLA-SLR (Ours) & 61.05 & 91.45 & 58.05 & 90.70 & 75.11 & \tbf{94.62} & 75.07 & \tbf{94.70} & 86.23 & \tbf{97.60} & 86.67 & \tbf{97.81} & 91.47 & \tbf{96.90} & 92.17 & \tbf{97.17} \\
NLA-SLR (Ours, 3-crop) & \tbf{61.26} & \tbf{91.77} & \tbf{58.31} & 90.91 & \tbf{75.64} & \tbf{94.62} & \tbf{75.72} & 94.65 & \tbf{86.98} & \tbf{97.60} & \tbf{87.33} & \tbf{97.81} & \tbf{92.64} & \tbf{96.90} & \tbf{93.08} & \tbf{97.17} \\

\bottomrule
\end{tabular}}
\vspace{-3mm}
\caption{Comparison with previous works on WLASL. The results of I3D and ST-GCN are reproduced by \cite{li2020word} and \cite{hu2021signbert}, respectively. BSL achieves multi-crop inference by sliding a window with a stride of 8 frames. ($\dagger$denotes methods using extra data. *denotes methods using much more modalities than ours, \eg, optical flow, depth map, and depth flow.)}
\label{tab:sota_wlasl}
\vspace{-5mm}
\end{table*}

\subsection{Datasets and Evaluation Metrics}
\noindent\textbf{Datasets.}
We evaluate our method on three public sign language recognition datasets: MSASL \cite{joze2019ms}, WLASL \cite{li2020word}, and NMFs-CSL \cite{hu2021global}.
\textit{MSASL} is an American sign language (ASL) dataset with a vocabulary size of 1,000. It consists of 16,054, 5,287, and 4,172 samples in the training, development (dev), and test set, respectively. It also released three subsets consisting of only the top 500/200/100 most frequent glosses.
\textit{WLASL} is the latest ASL dataset with a larger vocabulary size of 2,000. It consists of 14,289, 3,916, and 2,878 samples in the training, dev, and test set, respectively. Similar to MSASL, it also released three subsets consisting of 1,000/300/100 frequent glosses.
\textit{NMFs-CSL} is a challenging Chinese sign language (CSL) dataset involving many fine-grained non-manual features (NMFs). It consists of 25,608 and 6,402 samples in the training and test set with a vocabulary size of 1,067. However, since the dataset owners only provide label indexes instead of glosses, we cannot apply inter-modality mixup on it, and we have to replace our language-aware label smoothing with the vanilla one.

\noindent\tbf{Evaluation Metrics.}
Following \cite{hu2021signbert, hu2021hand, jiang2021sign}, we report both per-instance and per-class accuracy, which denote the average accuracy over instances and classes, on the test sets. Note that since NMFs-CSL is a balanced dataset, \ie, each class contains equal amount of samples, we only report per-instance accuracy on it.

\subsection{Implementation Details}
\noindent\tbf{Training Details and Hyper-parameters.}
The S3D backbone within VKNet-64/32 is first pretrained on Kinetics-400 \cite{kay2017kinetics}.
Then we separately pretrain the video and keypoint encoder within VKNet-64/32 on SLR datasets.
Finally, our VKNet is initialized with the pretrained VKNet-64 and VKNet-32.
Data augmentations include spatial cropping with a range of [0.7-1.0] and temporal cropping.
We adopt identical data augmentations for both RGB videos and heatmap sequences to maintain spatial and temporal consistency.
Unless otherwise specified, we set $\lambda \sim Beta(0.8, 0.8)$ for intra-modality mixup \cite{zhang2018mixup}, and $\epsilon=0.2$ and $\tau=0.5$ in Eq. \ref{equ:lang_lbsm}.
Similar to \cite{grill2020bootstrap}, we gradually increase $\mu$ in Eq. \ref{equ:ema} such that greater gradients of FC1 come from $\mathcal{L}_{CLS}$ in the late training stage since only FC1 is used during inference.
Specifically, $\mu=1-(1-\mu_{base})\cdot(\cos{(\pi m/M)}+1)/2$, where $\mu_{base}=0.99$, $m$ is the current epoch, and $M$ is the maximum number of epochs.
For the same reason, we gradually decrease the weight of $\mathcal{L}_{IMM}$ by $\gamma=(\cos{(\pi m/M)}+1)/2$.
The whole model is trained with a batch size of 32 for 100 epochs.
We use a cosine annealing schedule and an Adam optimizer \cite{adam} with a weight decay of $1e-3$ and an initial learning rate of $1e-3$.

\noindent\tbf{Inference.}
We report results of single-crop and 3-crop inference for a comparison with state-of-the-art methods \cite{albanie2020bsl, jiang2021skeleton, jiang2021sign}. All ablation studies are conducted in the setting of single-crop inference. For 3-crop inference, we temporally crop videos at the start, middle, end of the raw video, and the average prediction is served as the final prediction.
More details are in the supplementary materials.

\subsection{Comparison with State-of-the-art Methods}
\begin{table}[t]
\centering
\resizebox{0.75\linewidth}{!}{
\begin{tabular}{l|cc}
\toprule
Method & Top-1 & Top-5 \\

\midrule
I3D$^\Diamond$ \cite{I3D} & 64.4 & 88.0 \\
TSM$^\Diamond$ \cite{lin2019tsm} & 64.5 & 88.7 \\
Slowfast$^\Diamond$ \cite{feichtenhofer2019slowfast} & 66.3 & 86.6 \\
GLE-Net \cite{hu2021global} & 69.0 & 88.1 \\
HMA \cite{hu2021hand} & 75.6 & 95.3 \\
SignBERT$\dagger$ \cite{hu2021signbert} & 78.4 & 97.3 \\
BEST \cite{best} & 79.2 & 97.1 \\
\midrule

NLA-SLR (Ours) & 83.4 & 98.3 \\
NLA-SLR (Ours, 3-crop) & \tbf{83.7} & \tbf{98.5} \\

\bottomrule
\end{tabular}
}
\vspace{-3mm}
\caption{Comparison with previous works on NMFs-CSL. ($^\Diamond$methods reproduced by GLE-Net. $\dagger$methods using extra data.)}
\label{tab:sota_nmf}
\vspace{-5mm}
\end{table}

\noindent\tbf{MSASL.}
Table \ref{tab:sota_msasl} shows a comprehensive comparison between other methods and ours on all the sub-splits of MSASL.
Our approach outperforms the previous best method SignBERT \cite{hu2021signbert}, which utilizes extra data, by 2.56\%/2.50\%/1.46\% on the 1,000/200/100 sub-splits regarding the top-1 accuracy, respectively.

\noindent\tbf{WLASL.}
We evaluate our method on all the sub-splits of WLASL as shown in Table \ref{tab:sota_wlasl}.
The previous state-of-the-art method, SAM-SLR-v2 \cite{jiang2021sign}, proposes a heavy multi-modal ensemble framework, which involves many modalities including RGB videos, keypoints, optical flow, depth map, and depth flow.
However, our method significantly outperforms SAM-SLR-v2 by 1.87\%/1.68\% in terms of the per-instance/class top-1 accuracy while using much fewer modalities (only RGB videos and keypoints).

\noindent\tbf{NMFs-CSL.}
Finally, as shown in Table \ref{tab:sota_nmf}, our approach also outperforms the previous best method BEST \cite{best} by a large margin (83.7\% \vs 79.2\% on top-1 accuracy).

\subsection{Ablation Studies}
We conduct ablation studies on WLASL following \cite{li2020transferring, jiang2021sign} due to its large vocabulary size.

\noindent\tbf{VKNet.}
We first validate the effectiveness of our backbone, VKNet. As shown in Table \ref{tab:abl_vknet}, two-stream models, VKNet-32/64, can significantly outperform single-stream models, Video/Keypoint-32/64, which validates the effectiveness of modeling both videos and keypoints.
Besides, 64-frame models can consistently outperform 32-frame ones as expected since longer inputs can provide more information for the model to classify sign videos.
However, our VKNet performs better than a single 64-frame model, VKNet-64, especially on the top-5 accuracy, which implies that the 64-frame and 32-frame inputs can complement each other and the difference of the temporal receptive fields can bring more knowledge to model training.

\begin{table}[t]
\centering
\resizebox{0.8\linewidth}{!}{
\begin{tabular}{l|cc|cc}
\toprule
\multirow{2}{*}{Method} & \multicolumn{2}{c|}{Per-instance} & \multicolumn{2}{c}{Per-class} \\
& Top-1 & Top-5 & Top-1 & Top-5 \\

\midrule
Video-32 & 45.73 & 81.10 & 42.69 & 79.90\\
Keypoint-32 & 46.66 & 79.95 & 43.81 & 78.49\\
VKNet-32 & 52.95 & 85.75 & 50.26 & 84.50\\
\midrule
Video-64 & 51.15 & 83.43 & 48.14 & 82.20\\
Keypoint-64 & 49.10 & 82.00 & 46.18 & 80.71\\
VKNet-64 & 56.95 & 87.00 & 54.13 & 86.05\\
\midrule
VKNet & \tbf{57.19} & \tbf{88.29} & \tbf{54.35} & \tbf{87.49}\\
\bottomrule
\end{tabular}
}
\vspace{-3mm}
\caption{Ablation studies on VKNet.}
\label{tab:abl_vknet}
\vspace{-5mm}
\end{table}

\noindent\tbf{Major Components of NLA-SLR.}
As shown in Table \ref{tab:abl_main}, we study the effects of the major components of our NLA-SLR framework: language-aware label smoothing (Lang-LS) and sign mixup (ensemble of the intra- and inter-modality mixup).
First, Lang-LS can improve the performance of the baseline, VKNet, by 1.22\%/1.11\% on the top-1 and top-5 accuracy, respectively, which validates the effectiveness of language-aware soft labels.
Besides, more performance gain comes from sign mixup, which significantly improves the top-1 accuracy from 57.19\% to 60.32\%.
Finally, using both Lang-LS and sign mixup along with the VKNet can achieve the best performance: 61.05\%/91.45\% on the top-1 and top-5 accuracy, respectively.
Note that both of the major components introduce negligible extra cost: Lang-LS simply replace the one-hot labels with the language-aware soft labels; sign mixup merely introduces two extra fully-connected layers (one for mapping gloss features and the other one for auxiliary training) for each head network, and both of them are dropped during inference.

\begin{table}[t]
\setlength\tabcolsep{4pt}
\centering
\resizebox{\linewidth}{!}{
\begin{tabular}{c|cc|cc|cc}
\toprule
\multirow{2}{*}{VKNet} & \multirow{2}{*}{Lang-LS} & \multirow{2}{*}{Sign Mixup} & \multicolumn{2}{c|}{Per-instance} & \multicolumn{2}{c}{Per-class} \\
& & & Top-1 & Top-5 & Top-1 & Top-5 \\

\midrule
\checkmark & & & 57.19 & 88.29 & 54.35 & 87.49 \\
\checkmark & \checkmark & & 58.41 & 89.40 & 55.74 & 88.67 \\
\checkmark & & \checkmark & 60.32 & 90.86 & 57.55 & 90.06 \\
\checkmark & \checkmark & \checkmark & \tbf{61.05} & \tbf{91.45} & \tbf{58.05} & \tbf{90.70} \\

\bottomrule
\end{tabular}
}
\vspace{-3mm}
\caption{Ablation studies for the major components of NLA-SLR. (Lang-LS: language-aware label smoothing.)}
\label{tab:abl_main}
\vspace{-3mm}
\end{table}

\begin{table}[t]
\setlength\tabcolsep{4pt}
\centering
\resizebox{\linewidth}{!}{
\begin{tabular}{cc|cc|cc}
\toprule
\multicolumn{2}{c|}{Sign Mixup} & \multicolumn{2}{c|}{Per-instance} & \multicolumn{2}{c}{Per-class} \\
Intra-Modality & Inter-Modality & Top-1 & Top-5 & Top-1 & Top-5 \\

\midrule
& & 58.41 & 89.40 & 55.74 & 88.67 \\
\checkmark & & 59.56 & 90.10 & 56.77 & 89.33 \\
& \checkmark & 59.66 & 90.10 & 56.72 & 89.20 \\
\checkmark & \checkmark & \tbf{61.05} & \tbf{91.45} & \tbf{58.05} & \tbf{90.70} \\

\bottomrule
\end{tabular}
}
\vspace{-3mm}
\caption{Ablation studies on sign mixup which is composed of intra-modality and inter-modality mixup.}
\label{tab:abl_signmixup}
\vspace{-3mm}
\end{table}

\noindent\textbf{Sign Mixup.}
Our sign mixup is composed of two parts: intra-modality mixup, which extends the vanilla mixup \cite{zhang2018mixup} to keypoint heatmaps, and inter-modality mixup, which aims to maximize the signs' separability with the help of language information.
As shown in Table \ref{tab:abl_signmixup}, either intra- or inter-modality mixup can improve the performance by more than 1\% on the top-1 accuracy.
In addition, intra- and inter-modality mixup are compatible—using both mixup techniques surpasses using either one of them.

\begin{table}[t]
\setlength\tabcolsep{4pt}
\centering
\resizebox{\linewidth}{!}{
\begin{tabular}{ccc|cc|cc}
\toprule
Auxiliary & Inte- & Loss & \multicolumn{2}{c|}{Per-instance} & \multicolumn{2}{c}{Per-class} \\
Classifier & gration & Weight Decay & Top-1 & Top-5 & Top-1 & Top-5 \\

\midrule
& & & 59.56 & 90.10 & 56.77 & 89.33 \\
\checkmark & & & 59.87 & 90.31 & 57.07 & 89.57 \\
\checkmark & \checkmark & & 60.84 & 91.07 & 57.99 & 90.28 \\
\checkmark & \checkmark & \checkmark & \tbf{61.05} & \tbf{91.45} & \tbf{58.05} & \tbf{90.70} \\

\bottomrule
\end{tabular}
}
\vspace{-3mm}
\caption{Ablation studies for inter-modality mixup.}
\label{tab:abl_lang}
\vspace{-5mm}
\end{table}

\noindent\tbf{Inter-Modality Mixup.}
As shown in Table \ref{tab:abl_lang}, we first study the effects of the auxiliary classifier, FC2 in Figure \ref{fig:vl_mixup}.
It only slightly improves the performance (0.31\% on top-1 accuracy).
Most performance gain (almost 1\% on the top-1 accuracy) comes from the integration of the two classifiers (FC1 and FC2 as described in Section~\ref{sec:vl_mixup}). The reason is that it enables the natural language information to propagate from FC2 to FC1, which is the primary classifier during inference.
Finally, the loss weight decay strategy of $\mathcal{L}_{IMM}$ also has a positive effect since it assures that more gradients for FC1 come from $\mathcal{L}_{CLS}$ in the late training stage.

\noindent\tbf{Language-aware Label Smoothing.}
We conduct a comprehensive comparison between the vanilla label smoothing and our language-aware label smoothing (Lang-LS) by varying the smoothing parameter $\epsilon$ from 0.1 to 0.3.
As shown in Table \ref{tab:abl_lbsm}, our Lang-LS consistently outperforms the vanilla one regardless of the value of $\epsilon$.
The results suggest that for SLR models, assigning biased smoothing weights to the soft labels on the basis of gloss feature similarities (Eq. \ref{equ:lang_lbsm}) is a stronger regularization technique than the uniform distribution in the vanilla label smoothing (Eq. \ref{equ:vani_lbsm}).

\begin{table}[t]
\centering
\resizebox{0.8\linewidth}{!}{
\begin{tabular}{c|c|cc|cc}
\toprule
\multirow{2}{*}{$\epsilon$} & \multirow{2}{*}{Type} & \multicolumn{2}{c|}{Per-instance} & \multicolumn{2}{c}{Per-class} \\
& & Top-1 & Top-5 & Top-1 & Top-5 \\

\midrule
\multirow{2}{*}{0.1} & Vanilla & 59.83 & 90.72 & 56.90 & 90.10 \\
& Language & 60.15 & 91.35 & 57.30 & 90.68 \\
\midrule
\multirow{2}{*}{0.2} & Vanilla & 60.11 & 91.00 & 57.09 & 90.34 \\
& Language & \tbf{61.05} & \tbf{91.45} & \tbf{58.05} & \tbf{90.70} \\
\midrule
\multirow{2}{*}{0.3} & Vanilla & 60.01 & 90.97 & 57.01 & 90.12 \\
& Language & 60.49 & 91.31 & 57.44 & 90.67 \\

\bottomrule
\end{tabular}
}
\vspace{-3mm}
\caption{Comparison between the vanilla and language-aware label smoothing.}
\label{tab:abl_lbsm}
\vspace{-4mm}
\end{table}

\begin{table}[t]
\centering
\resizebox{\linewidth}{!}{
\begin{tabular}{l|lc|c|c}
\toprule
Method & VS-S & VS-D & Non-VS & Overall \\

\midrule
VKNet & 50.50 & 48.13 & 59.13 & 57.19 \\
+Lang-LS & 64.36 & 50.93 & 59.51 & 58.41 \\
+Lang-LS, Inter-Mixup & \textbf{65.35} & \textbf{56.07} & \textbf{60.07} & \textbf{59.66} \\

\bottomrule
\end{tabular}
}
\vspace{-3mm}
\caption{Quantitative results over VISigns. We report top-1 accuracy on WLASL2000. 
(VS-S/D: VISigns with similar/distinct semantic meanings.)
}
\label{tab:visign}
\vspace{-5mm}
\end{table}

\noindent \textbf{Presence and Quantitative Results of VISigns.}
To identify the VISigns appeared in the testing set, we first use our baseline model, VKNet, to get the highest prediction score $p_1$ (classified as gloss $g_1$) and the second highest prediction score $p_2$ (classified as gloss $g_2$) for each sample. Then we calculate the difference $\delta = p_1 - p_2$. If $\delta \leq 0.1$, we regard $g_1$ and $g_2$ as potential VISigns. Next, we calculate the gloss similarity $s$ of $g_1$ and $g_2$ via FastText. If $s\ge0.5$, we consider $g_1$ and $g_2$ as VS-S, otherwise, they are considered as VS-D. Finally, we invite native signers to filter out wrong cases. As a result, for WLASL with a vocabulary size of 2000, we get 101 instances covering 64 VS-S, 428 instances covering 270 VS-D, and 2349 instances covering 1666 non-VISigns (non-VS), respectively. As shown in Table \ref{tab:visign}, Lang-LS and Inter-Mixup yield the highest performance gains for VS-S (50.50 $\rightarrow$ 64.36) and VS-D (50.93 $\rightarrow$ 56.07), respectively, demonstrating that the improvements of our method derive from handling VISigns.
\vspace{-2mm}
\section{Conclusion}
\vspace{-1mm}
In this work, we propose Natural Language-Assisted Sign Language Recognition (NLA-SLR) framework, which leverages semantic information contained in glosses to promote sign language recognition.
Specifically, we first propose language-aware label smoothing to ease model training by generating soft labels whose smoothing
weights are the normalized semantic similarities.
Second, to maximize the separability of signs with distinct semantic meanings, we propose inter-modality mixup which blends vision and gloss features as well as their labels.
Besides, we also present a novel backbone, video-keypoint network, to model both RGB videos and human body keypoints and to absorb knowledge from sign videos of different temporal receptive fields.
Empirically, our approach surpasses previous best methods on three widely-adopted benchmarks.

\small\noindent \textbf{Acknowledgements.}
The work described in this paper was partially supported by a grant from the Research Grants Council of the HKSAR, China (Project No. HKUST16200118).

{\small
\bibliographystyle{ieee_fullname}
\bibliography{egbib}
}

\appendix
\section{More Implementation Details}
\begin{figure*}[t]
    \centering
    \begin{subfigure}[b]{0.45\textwidth}
     \centering
     \includegraphics[width=0.8\textwidth]{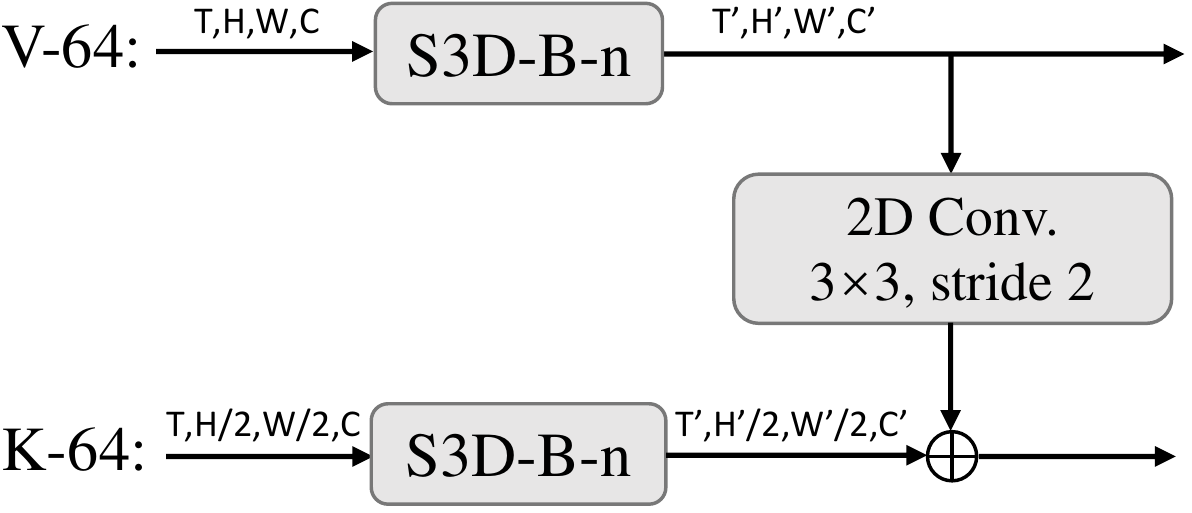}
     \caption{Lateral connection from the 64-frame video encoder to the 64-frame keypoint encoder. We use a 2D convolution layer with a kernel size of $3\times 3$ and a stride of 2 to match the spatial resolutions. The same holds true for the 32-frame input.}
     \label{fig:bilat_A}
    \end{subfigure}
    \quad\quad
    \begin{subfigure}[b]{0.45\textwidth}
     \centering
     \includegraphics[width=0.8\textwidth]{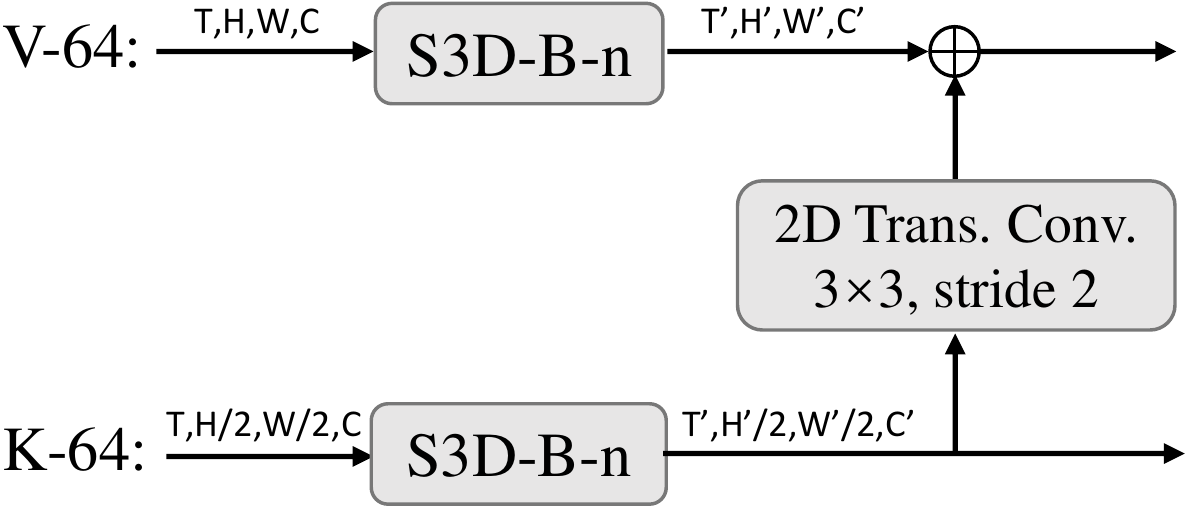}
     \caption{Lateral connection from the 64-frame keypoint encoder to the 64-frame video encoder. We use a 2D transposed convolution layer with a kernel size of $3\times 3$ and a stride of 2 to match the spatial resolutions. The same holds true for the 32-frame input.}
     \label{fig:bilat_B}
    \end{subfigure}
    
    \begin{subfigure}[b]{0.45\textwidth}
     \centering
     \includegraphics[width=0.8\textwidth]{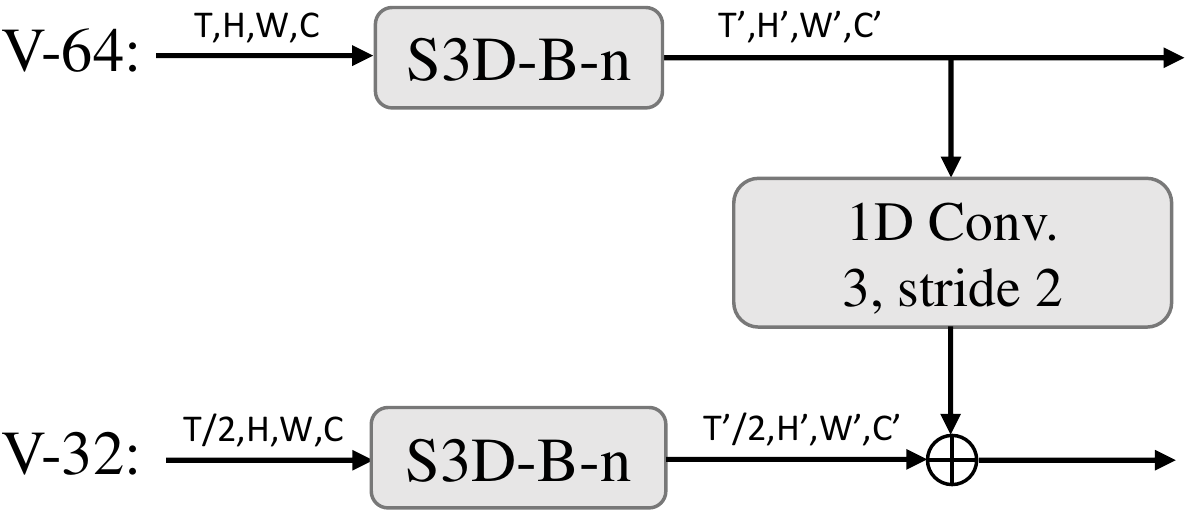}
     \caption{Lateral connection from the 64-frame video encoder to the 32-frame one. We use a 1D convolution layer with a kernel size of 3 and a stride of 2 to match the temporal resolutions. The same holds true for the keypoint encoder.}
     \label{fig:bilat_C}
    \end{subfigure}
    \quad\quad
    \begin{subfigure}[b]{0.45\textwidth}
     \centering
     \includegraphics[width=0.8\textwidth]{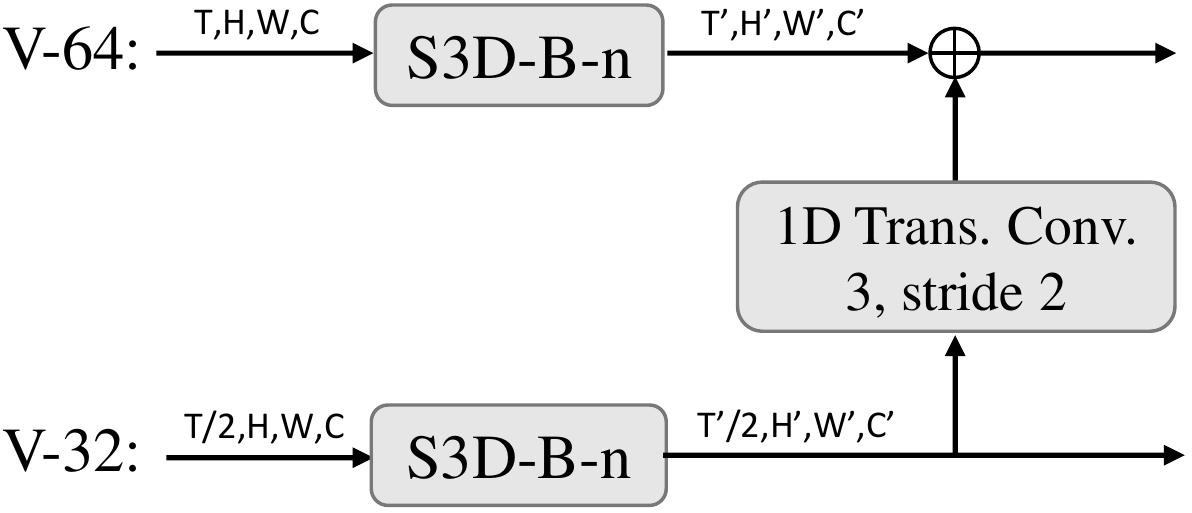}
     \caption{Lateral connection from the 32-frame video encoder to the 64-frame one. We use a 1D transposed convolution layer with a kernel size of 3 and a stride of 2 to match the temporal resolutions. The same holds true for the keypoint encoder.}
     \label{fig:bilat_D}
    \end{subfigure}

    \caption{Illustration of the lateral connections. Note that we split bidirectional lateral connections into unidirectional ones for better illustration.}
    \label{fig:bilat}
\end{figure*}

\noindent\textbf{Bidirectional Lateral Connections.}
We apply bidirectional lateral connections \cite{duan2022revisiting} to the first four S3D blocks for video-video, keypoint-keypoint, and video-keypoint information exchange.
For video-keypoint connections (dashed lines in Figure \textcolor{red}{4} in the main paper), since the input spatial resolutions of the video and keypoint encoder are $224\times224$ and $112\times112$, respectively, we use 2D convolution (Figure \ref{fig:bilat_A}) and transposed convolution layers (Figure \ref{fig:bilat_B}) with a stride of 2 and a kernel size of $3\times3$ to match the spatial resolutions.
For video-video and keypoint-keypoint connections (dotted dashed lines in Figure \textcolor{red}{4} in the main paper), due to the input length difference, we use 1D convolution (Figure \ref{fig:bilat_C}) and transposed convolution layers (Figure \ref{fig:bilat_D}) with a stride of 2 and a kernel size of 3 to match the temporal resolutions. Figure~\ref{fig:bilat} shows the bidirectional lateral connections.

\begin{figure*}[t]
\centering
\includegraphics[width=0.7\linewidth]{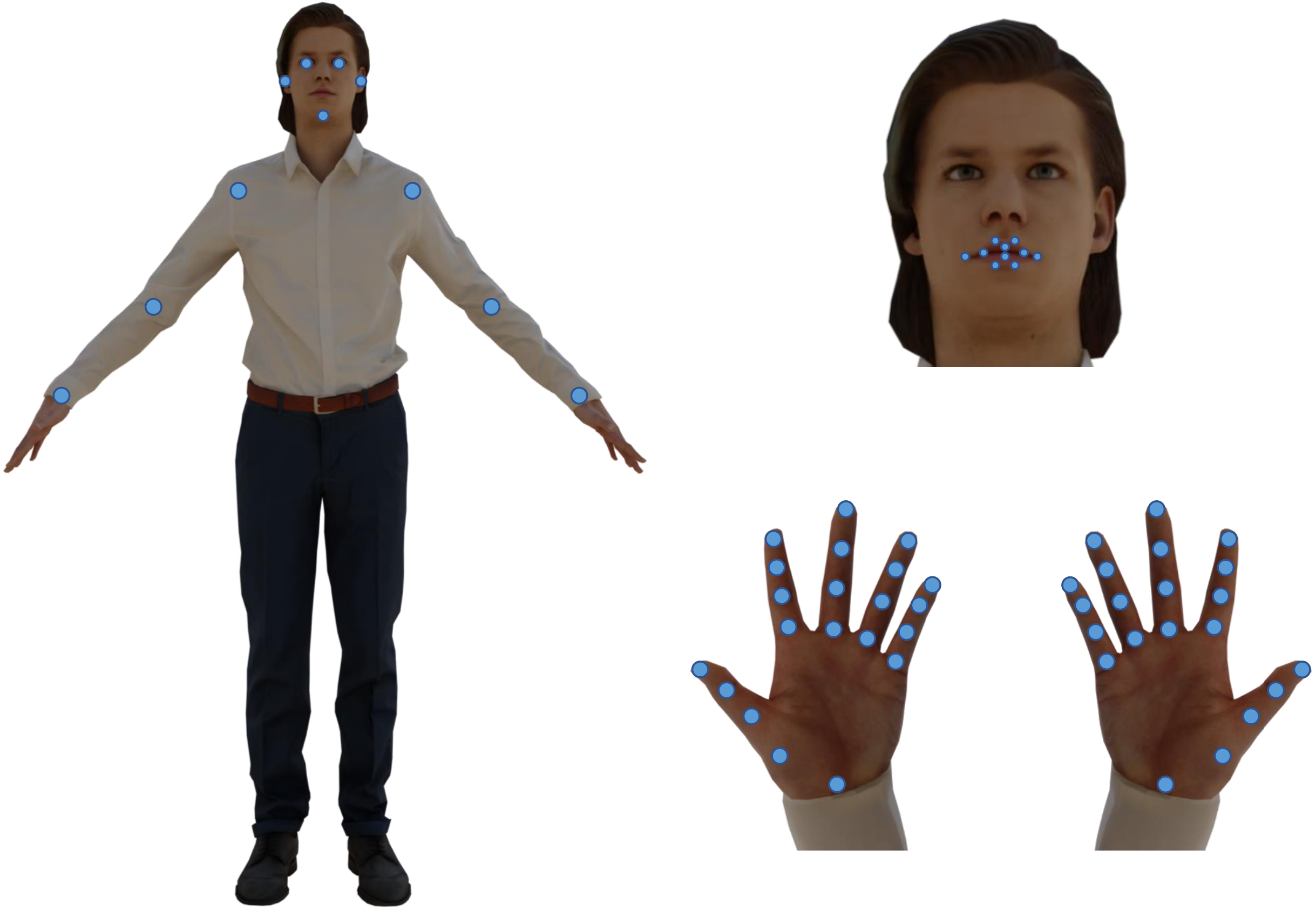}
\caption{Illustration of the keypoints (11 upper body keypoints, 10 mouth keypoints, and 42 hand keypoints) used in our VKNet.}
\label{fig:kp_illustration}
\end{figure*}

\noindent\textbf{Keypoint Illustration.} We show the keypoints used in our VKNet in Figure \ref{fig:kp_illustration}. The keypoints are estimated by HRNet~\cite{sun2019deep} trained on COCO-WholeBody~\cite{jin2020whole}.
We use a subset of keypoints including 11 upper body keypoints, 10 mouth keypoints, and 42 hand keypoints.

\section{More Experiments}
\noindent\tbf{Head Choices for Inter-Modality Mixup.}
By default, we apply our inter-modality mixup on all head networks.
To validate the effectiveness of this setting, we further conduct experiments on only applying it on partial heads.
We categorize the head networks into three groups: video heads with input features $(\boldsymbol{f}_{64}^V, \boldsymbol{f}_{32}^V)$, keypoint heads with input features $(\boldsymbol{f}_{64}^K, \boldsymbol{f}_{32}^K)$, and joint heads with input features $(\boldsymbol{f}_{64}, \boldsymbol{f}_{32}, \boldsymbol{f})$. See Figure \textcolor{red}{4} in the main paper for their definitions.
Table \ref{tab:abl_head} shows that applying the inter-modality mixup on either one group of heads outperforms the baseline, and our default setting, applying the inter-modality mixup on all heads, achieves the best performance.

\begin{table}[t]
\centering
\resizebox{\linewidth}{!}{
\begin{tabular}{ccc|cc|cc}
\toprule
\multirow{2}{*}{Video} & \multirow{2}{*}{Keypoint} & \multirow{2}{*}{Joint} & \multicolumn{2}{c|}{Per-instance} & \multicolumn{2}{c}{Per-class} \\
& & & Top-1 & Top-5 & Top-1 & Top-5 \\

\midrule
& & & 59.56 & 90.10 & 56.77 & 89.33 \\
\checkmark & & & 60.42 & 91.07 & 57.62 & 90.37 \\
& \checkmark & & 60.08 & 90.62 & 57.27 & 89.76 \\
& & \checkmark & 59.83 & 90.72 & 56.88 & 90.11 \\
\checkmark & \checkmark & & 60.56 & 91.24 & 57.87 & 90.37 \\
\checkmark & \checkmark & \checkmark & \tbf{61.05} & \tbf{91.45} & \tbf{58.05} & \tbf{90.70} \\
\bottomrule
\end{tabular}
}
\caption{Ablation studies on applying inter-modality mixup on different types of head networks.}
\label{tab:abl_head}
\end{table}

\begin{table}[ht]
\setlength\tabcolsep{3pt}
\centering
\resizebox{\linewidth}{!}{
\begin{tabular}{cccc|cc|cc}
\toprule
\multirow{2}{*}{Upper Body} & \multirow{2}{*}{Hand} & \multirow{2}{*}{Mouth} & \multirow{2}{*}{\#Keypoints} &  \multicolumn{2}{c|}{Per-instance} & \multicolumn{2}{c}{Per-class} \\
& & & & Top-1 & Top-5 & Top-1 & Top-5 \\

\midrule
\checkmark & & & 11 & 21.37 & 50.66 & 19.78 & 49.00 \\
\checkmark & \checkmark & & 53 & 48.54 & 81.45 & 45.52 & 79.94 \\
& \checkmark & \checkmark & 52 & 48.64 & 81.83 & 45.64 & 80.36 \\
\checkmark & \checkmark & \checkmark & 63 & \tbf{49.10} & \tbf{82.00} & \tbf{46.18} & \tbf{80.71}\\

\bottomrule
\end{tabular}
}
\caption{Ablation study on keypoint selection.}
\label{tab:abl_kp}
\end{table}

\noindent\textbf{Keypoint Selection.} We utilize HRNet \cite{sun2019deep} trained on COCO-WholeBody \cite{jin2020whole} to estimate 63 keypoints (11 for upper body, 42 for hands, and 10 for mouth) per frame.
As shown in Table \ref{tab:abl_kp}, we validate the effectiveness of each keypoint group by training several single-stream keypoint encoders.
Only using upper body keypoints yields the lowest top-1 accuracy (21.37\%).
Employing hand keypoints significantly improves the top-1 accuracy by 27.17\%.
This result is also consistent to the fact that sign languages mainly convey information by signers' hand movement.
Finally, the mouth keypoints also have a positive effect since signers usually mouth the words during signing.

\begin{table}[t]
\centering
\begin{tabular}{ccc|cc|cc}
\toprule
\multirow{2}{*}{V-V} & \multirow{2}{*}{K-K} & \multirow{2}{*}{V-K} & \multicolumn{2}{c|}{Per-instance} & \multicolumn{2}{c}{Per-class} \\
& & & Top-1 & Top-5 & Top-1 & Top-5 \\

\midrule
& & & 56.85 & 86.87 & 53.34 & 85.60 \\
\checkmark & & & 57.12 & 87.11 & 54.21 & 85.94 \\
\checkmark & \checkmark & & 57.16 & 87.56 & 54.03 & 86.54 \\
\checkmark & \checkmark & \checkmark & \tbf{57.19} & \tbf{88.29} & \tbf{54.35} & \tbf{87.49} \\
\bottomrule
\end{tabular}
\caption{Ablation studies on different types of bidirectional lateral connections. (V-V: video-video; K-K: keypoint-keypoint; V-K: video-keypoint.)}
\label{tab:abl_lat}
\end{table}

\noindent\textbf{Bidirectional Lateral Connections.} Within the VKNet, we apply bidirectional lateral connections \cite{duan2022revisiting} for video-video, keypoint-keypoint, and video-keypoint information exchange. See Figure \textcolor{red}{4} in the main paper for their illustration.
As shown in Table \ref{tab:abl_lat}, each type of bidirectional lateral connections has a positive effect on model performance, and our default setting, using all of the three types of the lateral connections, can achieve the best performance.

\begin{table}[t]
\centering
\begin{tabular}{l|cc|cc}
\toprule
\multirow{2}{*}{Method} &  \multicolumn{2}{c|}{Per-instance} & \multicolumn{2}{c}{Per-class} \\
& Top-1 & Top-5 & Top-1 & Top-5 \\

\midrule
SlowFast & 56.81 & 87.60 & 53.69 & 86.68 \\
VKNet & \tbf{57.19} & \tbf{88.29} & \tbf{54.35} & \tbf{87.49} \\

\bottomrule
\end{tabular}
\caption{Comparison between SlowFast and our VKNet.}
\label{tab:abl_slowfast}
\end{table}

\noindent\textbf{VKNet vs. SlowFast.} Our VKNet consists of two sub-networks, VKNet-64 and VKNet-32, to jointly model video-keypoint pairs with different temporal receptive fields.
The results in Table \textcolor{red}{4} in the main paper suggest that modeling different video-keypoint pairs with varied temporal receptive fields improves the model generalization capability. One network that is related to our VKNet is SlowFast \cite{feichtenhofer2019slowfast}, which consists of two streams taking RGB videos with low/high frame rate as inputs while having a fixed temporal receptive field.
For a fair comparison between SlowFast and our VKNet, we replace the ``temporal crop" operation in Figure \textcolor{red}{3} in the main paper with ``temporal sampling", \ie, sampling a 32-frame pair from the 64-frame one with a stride of 2 frames, to mimic the SlowFast.
As shown in Table \ref{tab:abl_slowfast}, our VKNet can consistently outperform SlowFast on all of the four metrics, showing that VKNet is a stronger backbone for sign language recognition.

\begin{table}[t]
\centering
\resizebox{\linewidth}{!}{
\begin{tabular}{l|cc|cc}
\toprule
\multirow{2}{*}{Method} &  \multicolumn{2}{c|}{Per-instance} & \multicolumn{2}{c}{Per-class} \\
& Top-1 & Top-5 & Top-1 & Top-5 \\

\midrule
Contrastive Learning & 59.90 & 91.28 & 57.23 & 90.59 \\
Inter-Modality Mixup & \tbf{61.05} & \tbf{91.45} & \tbf{58.05} & \tbf{90.70} \\

\bottomrule
\end{tabular}
}
\caption{Comparison between contrastive learning and our inter-modality mixup.}
\label{tab:abl_contras}
\end{table}

\noindent\textbf{Inter-Modality Mixup vs. Contrastive Learning.} Our inter-modality mixup blends vision and language features to better maximize the separability of signs.
Its effectiveness is shown in Table \textcolor{red}{6} in the main paper. One work that is related to our inter-modality mixup is CLIP \cite{clip}, which jointly trains an image encoder and a text encoder with a contrastive loss by maximizing the cosine similarity of positive image-text pairs while minimizing the similarity of negative pairs.
Following the practice in CLIP, we replace our inter-modality mixup loss $\mathcal{L}_{IMM}$ with a contrastive loss between the vision feature $\boldsymbol{f}$ and gloss features $\bar{\boldsymbol{E}}$.
As shown in Table \ref{tab:abl_contras}, our inter-modality mixup can consistently outperform the contrastive learning method on all of the four metrics. The results demonstrate that our inter-modality mixup is a more effective approach to exploit semantic information contained in glosses.

\begin{table}[t]
\centering
\begin{tabular}{l|cc|cc}
\toprule
\multirow{2}{*}{Method} &  \multicolumn{2}{c|}{Per-instance} & \multicolumn{2}{c}{Per-class} \\
& Top-1 & Top-5 & Top-1 & Top-5 \\

\midrule
Word2vec \cite{word2vec} & 60.63 & 91.14 & 57.53 & 90.42 \\
GloVe \cite{glove} & 60.81 & 90.90 & 57.73 & 90.27 \\
FastText \cite{mikolov2018advances} & \tbf{61.05} & \tbf{91.45} & \tbf{58.05} & \tbf{90.70} \\
BERT \cite{kenton2019bert} & 60.11 & 90.83 & 57.15 & 90.05 \\

\bottomrule
\end{tabular}
\caption{Comparison among different word representation learning methods.}
\label{tab:abl_word}
\end{table}

\noindent\textbf{Word Representation Learning Methods.} We adopt fastText \cite{mikolov2018advances} as our default gloss feature extractor. Here we investigate other alternatives as shown in Table~\ref{tab:abl_word}. Word2vec \cite{word2vec} and GloVe \cite{glove} are two classical word representation learning methods which are widely-adopted in NLP community. They perform comparably to each other that GloVe achieves better results on the top-1 accuracy while word2vec is superior regarding to the top-5 accuracy. As an improvement of word2vec, fastText leads to better results on all of the four metrics. Finally, we also utilize an advanced model, BERT-base \cite{kenton2019bert}, to extract word representations by averaging the outputs of the last layer. However, it performs worse than all the other methods since it is not dedicated to word representation learning.

\section{Visualization}
\begin{figure}[t]
\centering
\includegraphics[width=1.0\linewidth]{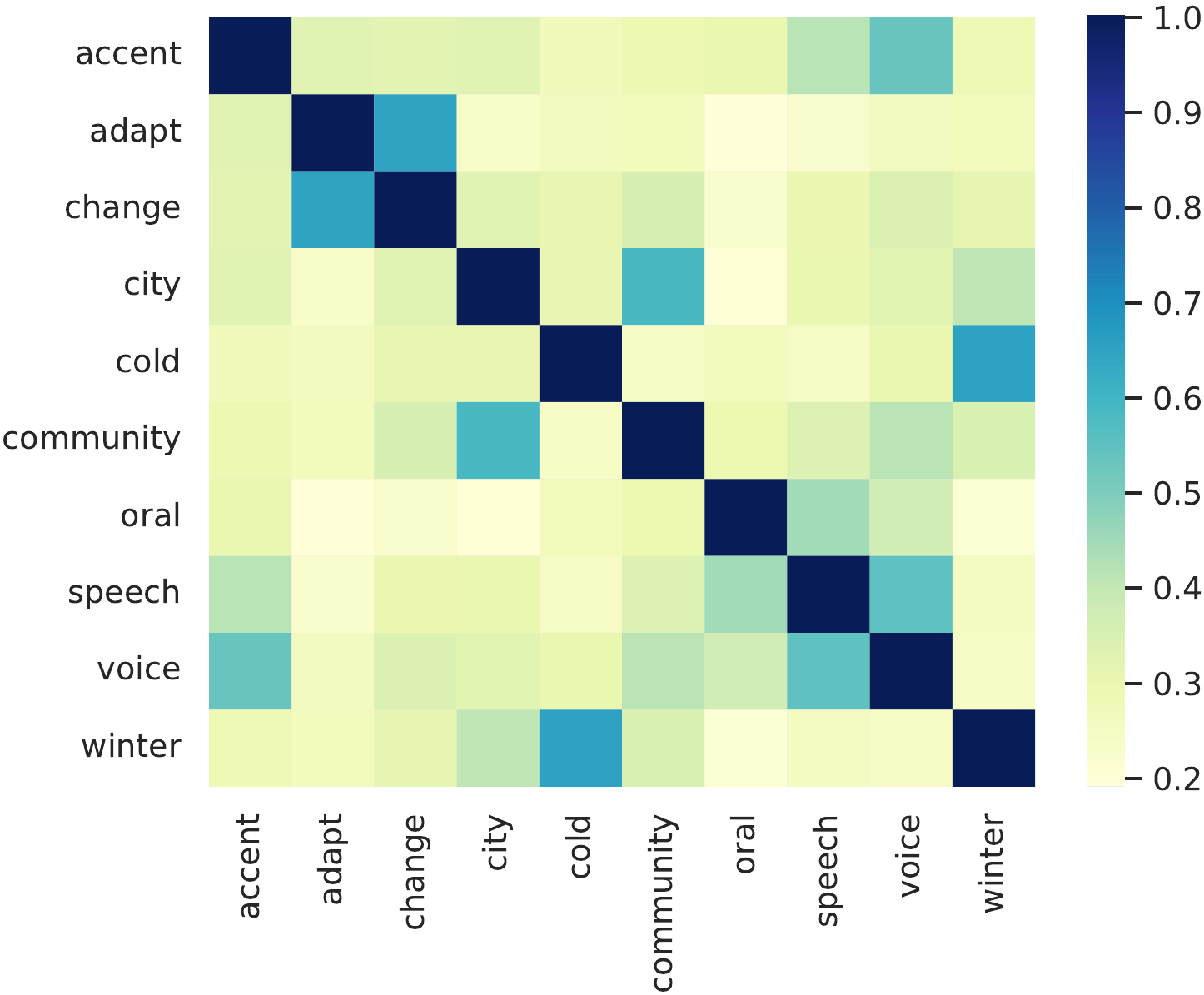}
\caption{Visualization of gloss feature similarities. We adopt fastText to extract gloss features.}
\label{fig:word_sim}
\end{figure}

\noindent\textbf{Gloss Feature Similarity.} The gloss feature similarities play a key role in our language-aware label smoothing.
We select several glosses from the vocabulary and visualize the cosine similarities between their gloss features as a heatmap in Figure \ref{fig:word_sim}.
We can see that the similarity matrix can roughly reflect the semantic similarities between glosses.
For example, the pairs: (``adapt", ``change"), (``city", ``community"), (``cold", ``winter"), (``speech", ``oral"), and (``accent", ``voice"), have high similarities, which are consistent to human understanding.

\begin{figure*}[t]
     \centering
     \begin{subfigure}[t]{0.95\textwidth}
         \centering
         \includegraphics[width=\textwidth]{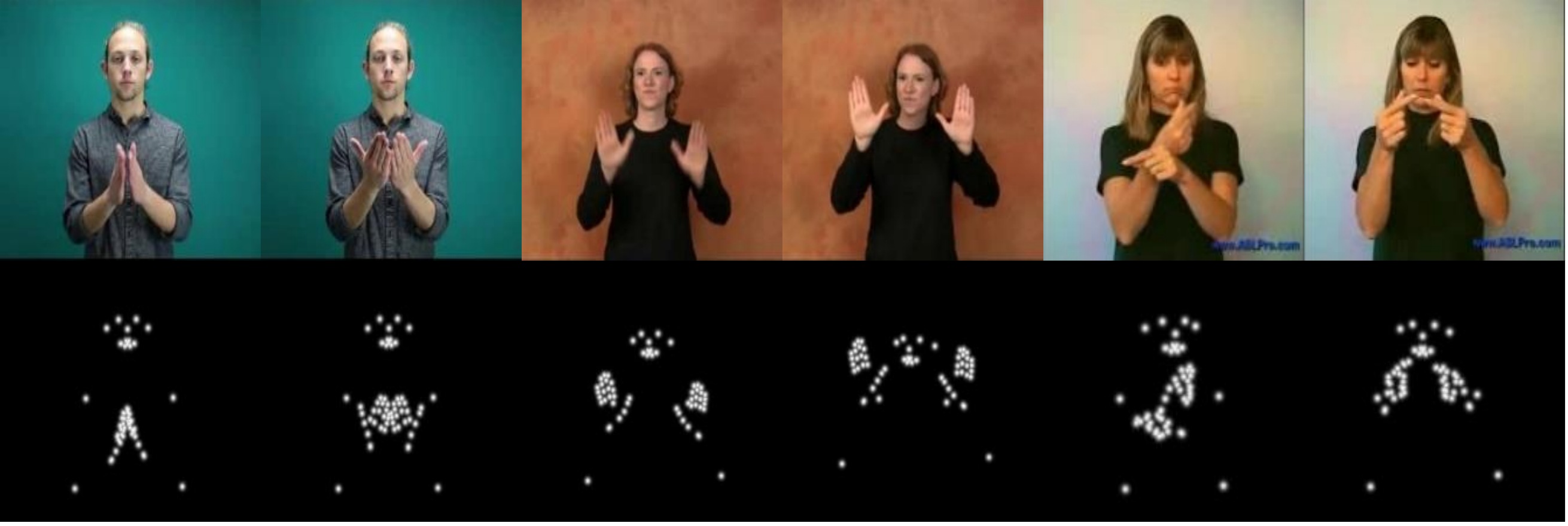}
         \caption{WLASL2000.}
         \label{fig:hmap_wlasl}
     \end{subfigure}

     \begin{subfigure}[t]{0.95\textwidth}
         \centering
         \includegraphics[width=\textwidth]{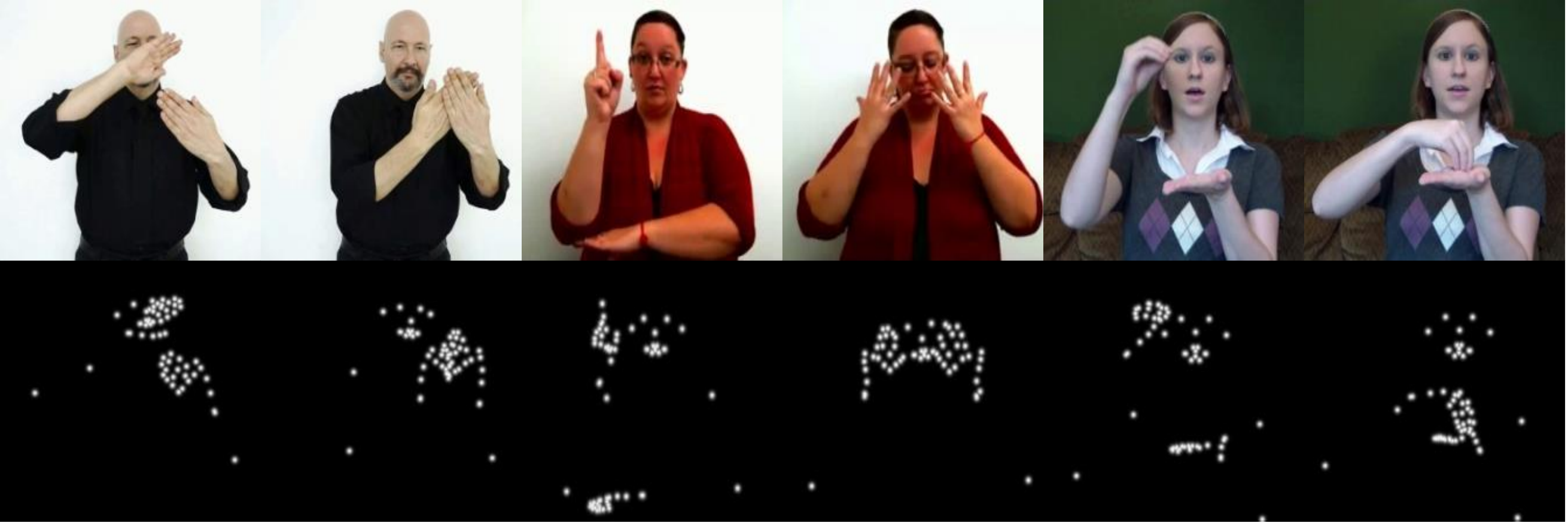}
         \caption{MSASL1000.}
         \label{fig:hmap_msasl}
     \end{subfigure}
     
     \begin{subfigure}[t]{0.95\textwidth}
         \centering
         \includegraphics[width=\textwidth]{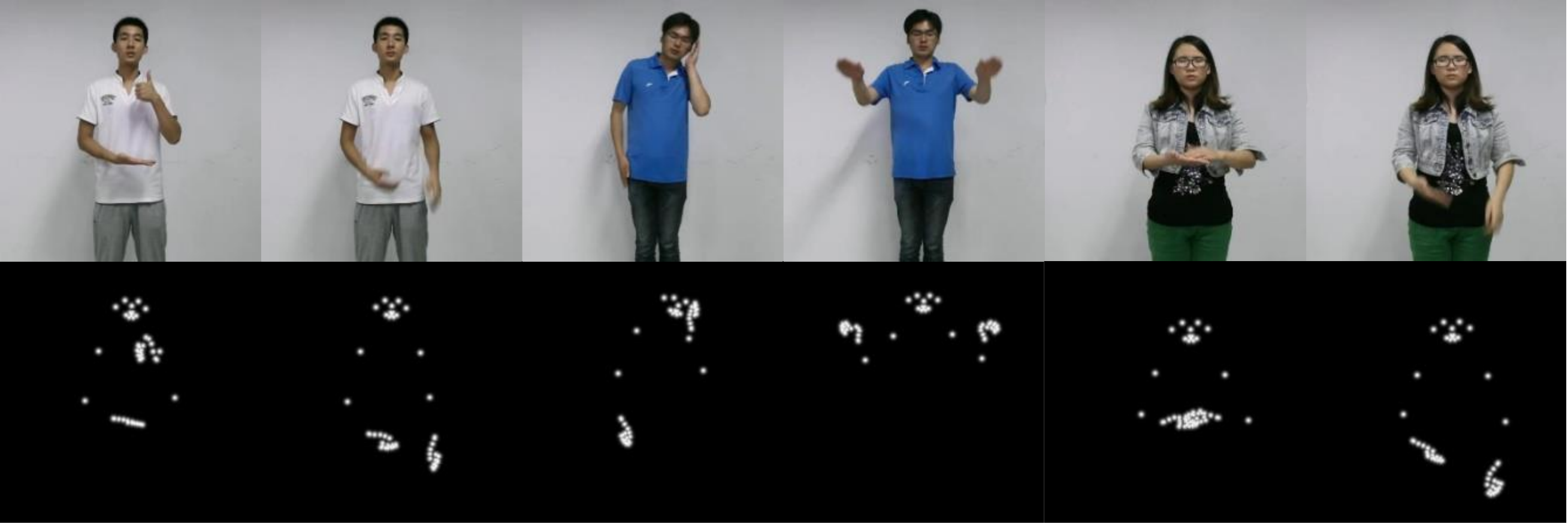}
         \caption{NMFs-CSL.}
         \label{fig:hmap_nmf}
     \end{subfigure}

\caption{Visualizations for the randomly selected frames and their corresponding keypoint heatmaps estimated by HRNet.}
\label{fig:hmap}
\end{figure*}

\noindent\textbf{Keypoint Heatmaps.} As shown in Figure \ref{fig:hmap}, we visualize the keypoint heatmaps extracted by HRNet \cite{sun2019deep} by randomly selecting six frames of three signers from the test sets of WLASL2000, MSASL1000, and NMFs-CSL, respectively.
We can clearly see that the heatmaps are robust to signer appearances, background variations, hand positions, and palm orientations.

\begin{figure*}[t]
     \centering
     \begin{subfigure}[t]{0.95\textwidth}
         \centering
         \includegraphics[width=\textwidth]{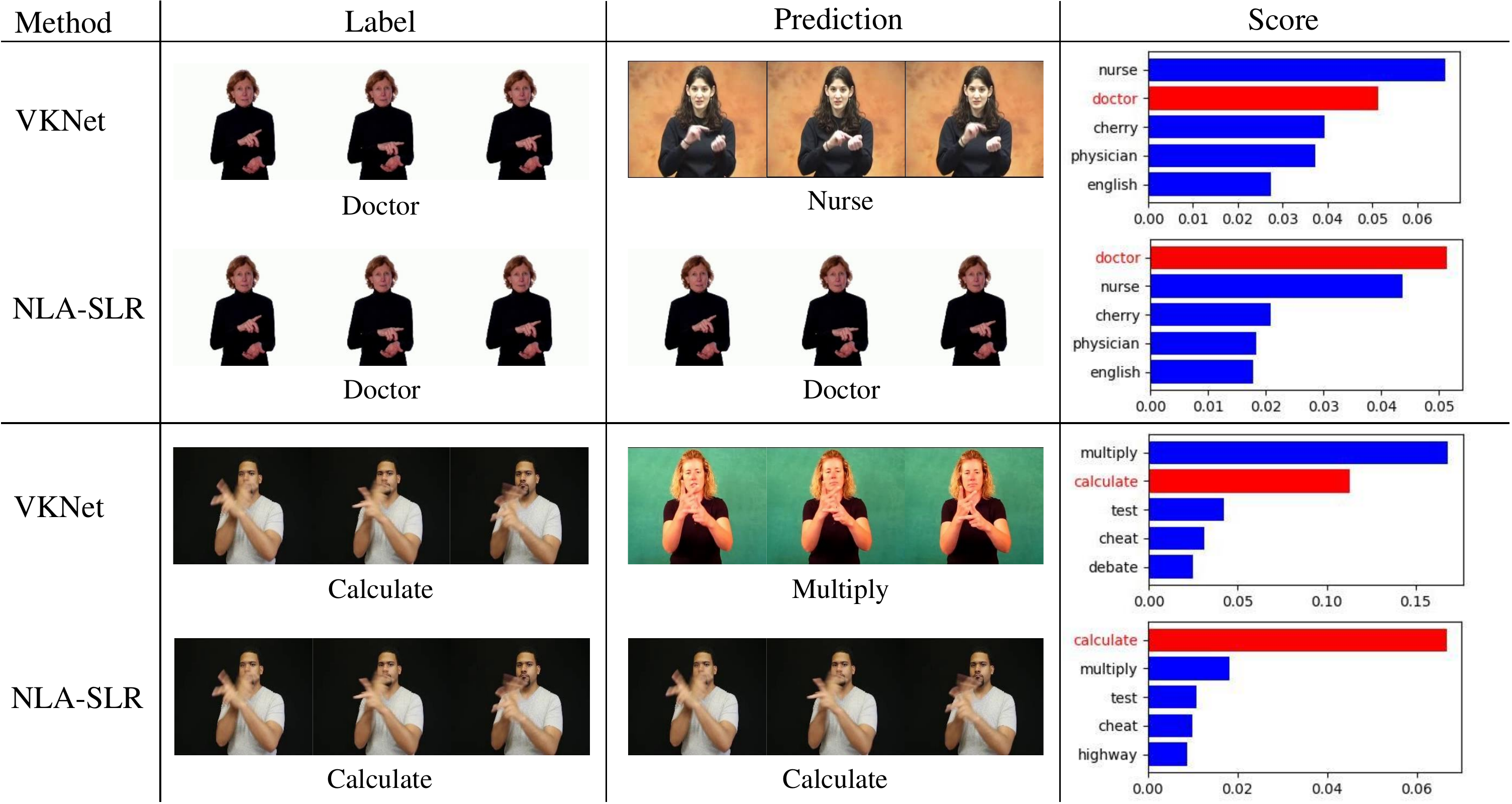}
         \caption{VISigns with \textit{similar} semantic meanings.}
         \label{fig:qual_res_A}
     \end{subfigure}

     \begin{subfigure}[t]{0.95\textwidth}
         \centering
         \includegraphics[width=\textwidth]{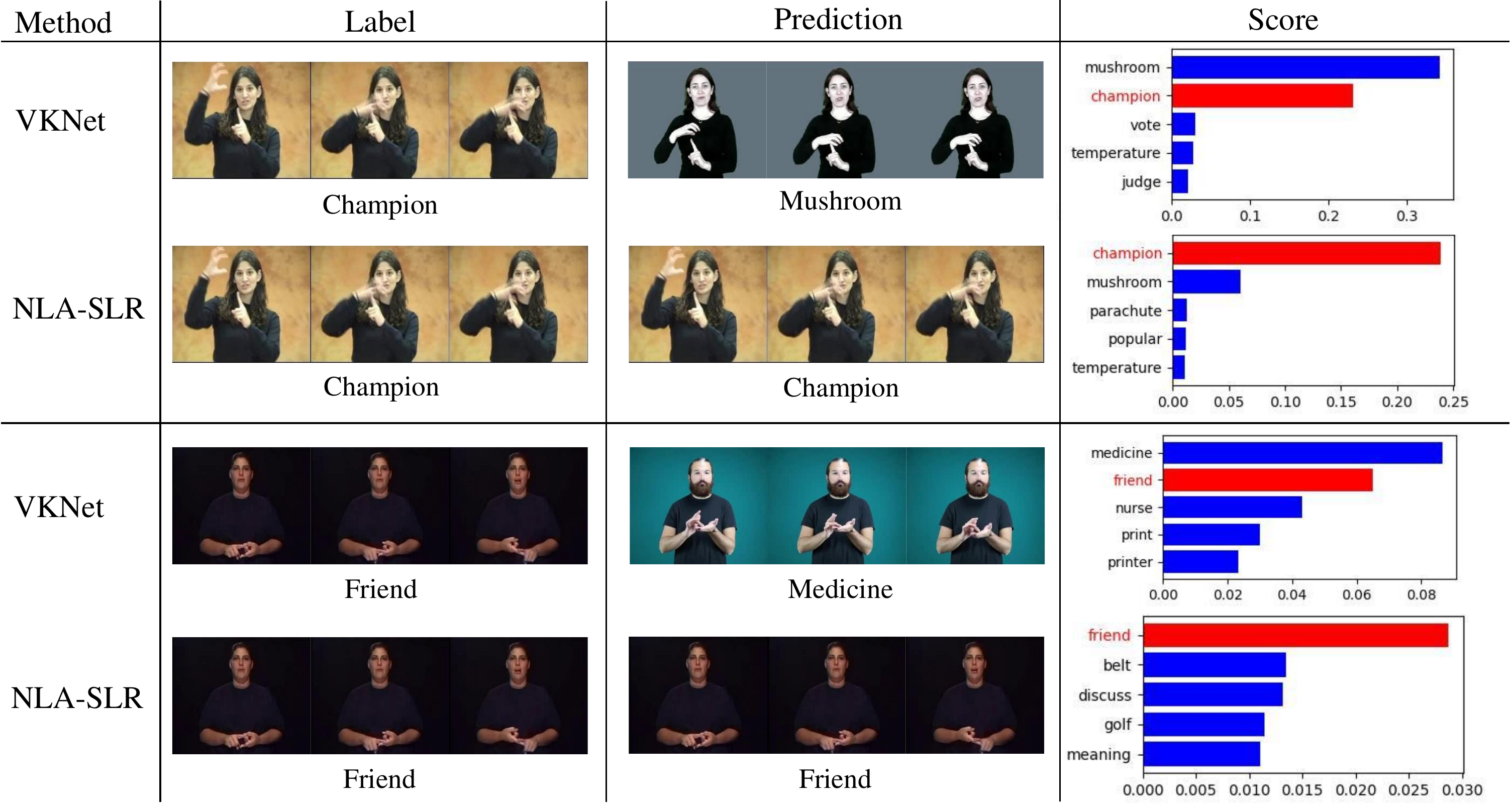}
         \caption{VISigns with \textit{distinct} semantic meanings.}
         \label{fig:qual_res_B}
     \end{subfigure}

\caption{Qualitative results on WLASL2000. (Here for NLA-SLR, we do not use intra-modality mixup for a fair comparison. The ground-truth gloss is highlighted in \textcolor{red}{red}.)}
\label{fig:qual_res}
\end{figure*}

\section{Qualitative Results}
As shown in Figure \ref{fig:qual_res}, we conduct qualitative analysis for our NLA-SLR.
We find that compared with VKNet (baseline), our NLA-SLR can well classify visually indistinguishable signs (VISigns) with either similar or distinct meanings. 
As shown in Figure \ref{fig:qual_res_A}, our NLA-SLR can successfully distinguish (``doctor", ``nurse") and (``calculate", ``multiply"), which are VISigns with similar semantic meanings, whereas the baseline, VKNet, fails to classify them.
Besides, as shown in Figure \ref{fig:qual_res_B}, our NLA-SLR can also recognize VISigns with distinct semantic meanings: (``champion", ``mushroom") and (``friend", ``medicine").
We owe these success to the two proposed techniques: language-aware label smoothing and inter-modality mixup.

\section{Social Impact and Limitation}
Sign language is the primary communication method among the deaf community.
Thus, research on sign language recognition can help bridge the communication gap between the normal-hearing and hearing-impaired people.

The proposed method is data-driven.
Thus, the model performance may be affected by the biases in the training data.
Besides, our backbone relies on pre-extracted keypoints; inaccurate keypoint estimation may hurt the model performance.
We believe that stronger keypoint estimators may further improve sign language recognition in the future.

\end{document}